\documentclass[runningheads]{llncs}
\usepackage{graphicx}
\usepackage{subfigure}
\usepackage{amsmath,amssymb} 
\usepackage{color}
\usepackage[width=122mm,left=12mm,paperwidth=146mm,height=193mm,top=12mm,paperheight=217mm]{geometry}
 \usepackage{upgreek}
 \usepackage[titletoc,title]{appendix}
 \usepackage{subfigure}
\begin{document}
\pagestyle{headings}
\mainmatter
\def\ECCV16SubNumber{1813}  

\title{Robust and Efficient Relative Pose with a Multi-camera System for Autonomous Vehicle in Highly Dynamic Environments} 

\titlerunning{Relative Pose Estimation for Autonomous Vehicle}

\authorrunning{Liu Liu, Hongdong Li and Yuchao Dai}

\author{Liu Liu \inst{1,2} \and Hongdong Li \inst{2,3} \and Yuchao Dai \inst{2} }

\institute{Northwestern Polytechnical University, China \and Australian National University, Australia \and Australia Centre of Excellence for Robotic Vision (ACRV), Australia 
}

\maketitle
\begin{abstract}
This paper studies the relative pose problem for autonomous vehicle driving in highly dynamic and possibly cluttered environments. This is a challenging scenario due to the existence of multiple, large, and independently moving objects in the environment, which often leads to excessive portion of outliers and results in erroneous motion estimation.  Existing algorithms cannot cope with such situations well.  This paper proposes a new algorithm for relative pose using a multi-camera system with multiple non-overlapping individual cameras. The method works robustly even when the numbers of outliers are overwhelming.  By exploiting specific prior knowledge of driving scene we have developed an efficient 4-point algorithm for multi-camera relative pose, which admits analytic solutions by solving a polynomial root-finding equation, and runs extremely fast (at about 0.5$\upmu$s per root).  When the solver is used in combination with RANSAC, we are able to quickly prune unpromising hypotheses, significantly improve the chance of finding inliers. Experiments on synthetic data have validated the performance of the proposed algorithm. Tests on real data further confirm the method's practical relevance.
\end{abstract} 
%
\section{Motivation}

Estimating the relative pose  between two camera views is a fundamental problem in 3D computer vision.  This problem has been extensively researched, and a large number of algorithms have been developed,  the most well known ones being the normalised 8-point algorithm \cite{Normalized-8-point:Hartley-1997} and the 5-point minimal algorithm \cite{Nister-Five-Point:PAMI-2004}.     

This paper is primarily concerned with relative pose in the context of autonomous vehicles. It is motivated by a practical desire to develop {\em highly robust } relative pose algorithm for estimating vehicle's ego-motion in highly dynamic environments with multiple independently moving objects.    Such a multi-body dynamic scenario,  despite  common in reality,  poses a significant challenge to relative pose problem.   Existing relative pose algorithms (e.g. 8-point and 5-point with RANSAC) cannot cope with this scenario well.   This is because, the existence of multiple, large, and independently moving objects often obscure the camera's field-of-view, resulting in excessive numbers of outliers in the scene.  When the  outlier ratio is significantly higher than 50\%, RANSAC becomes extremely inefficient, unsuitable for real-time tasks.  

\begin{figure}[h!] 
\centering
\includegraphics[width=0.35\textwidth]{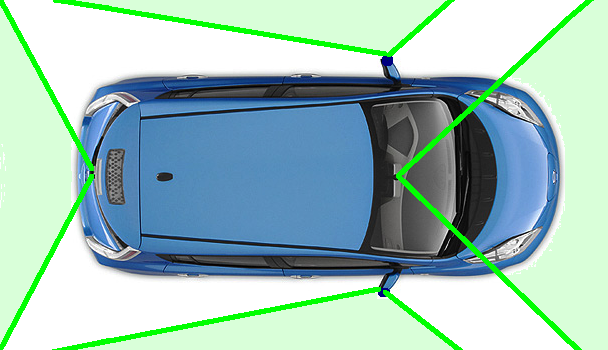}~~~~~~~~~~~~~\includegraphics[width=0.35\textwidth]{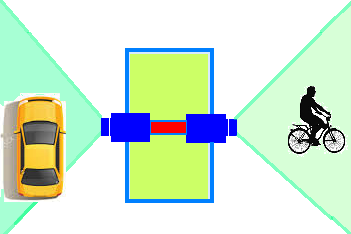}
\caption{\label{fig:config} {\small {\bf Left:} A typical multi-camera system configuration on a car.  These individual cameras do not share common field of views;  ~~ {\bf Right:}  The main idea of the paper for outlier removal is that,  although feature points on the moving objects may overwhelming in each individual camera's field-of-view,  their motion estimations can be quickly excluded during RANSAC computation,  because these motions are in general do not {\em compatible} with each other.   Details are given in the text.}} 
\end{figure}

To develop highly robust relative pose algorithm for autonomous driving in a dynamic multi-body environment is the primary goal of the current paper. Moreover, the paper focuses on the use of multi-camera rig system for robust relative pose. Multi-camera systems have received increasing attention in the area of autonomous vehicles \cite{Generalized-Camera-Motion-Lee:CVPR-2013}, because it is natural and convenient to mount multiple cameras, each facing a different direction, on a vehicle. For example, figure \ref{fig:config} shows a typical configuration of a multi-camera rig system on a car with two side-viewing, one front-viewing, and one rear-viewing cameras.   Often,  these individual cameras (of the multi-camera system) are arranged in a way so that they have minimum (or zero) overlapping field-of-views (FOVs), in order to give the vehicle a wider combined FOV coverage for better surrounding perception. Previous studies have shown that a wider FOV improves camera motion estimation accuracy \cite{Kim2010}. Having a wider FOV not only provides the benefit of better accuracy, but also offers an effective way to distinguish inliers from outliers; and our paper is capitalised on the second point.
Specifically, this paper proposes a new, simple and robust, multi-camera based relative pose method for autonomous road vehicle motion estimation. Our method is able to cope with highly dynamic environments having multi-body movements. It works even when most of the individual camera's field-of-view is obscured by multiple other independently moving objects.  The key idea of our method is simple and straightforward. It is based on a common observation, namely,  under daily driving situations, an independently moving object in the scene (such as a pedestrian, a bicycle, or other vehicles on road) can {\em rarely be simultaneously seen by two individual cameras}.  
 
There are many multi-camera relative pose algorithms existing in the literature (e.g., such as the linear 17-point algorithm \cite{Li07}, or the minimal 6-point solver \cite{Stewenius06}) , which can be applied for this purpose of relative pose in autonomous driving. They however suffer from severe computational complexity issue, therefore do not meet the real-time requirement for ego-motion computation in autonomous driving. For example, if it were to be used in conjunction with RANSAC with acceptable confidence, the 17-point algorithm would require massive number of random samples in the order of millions. On the other hand,  whilst the 6-point algorithm itself only requires to sample minimally 6 points, to solve it needs to build a complicated Groebner solver, which moreover can yield up to 64 solutions which must be further pruned, rendering the overall computations extremely expensive and not practical. By exploiting prior knowledges that are specific to road vehicle applications, this paper develops an extremely efficient minimal solver for multi-camera relative pose using only 4 feature points. By combining it with a new RASNAC sampling scheme, we achieve highly efficient and highly robust road vehicle ego-motion estimation.  
 \subsection{Contributions}
\begin{itemize}
\item  Our method is extremely fast, which runs at about 2.3 $\upmu$s  and generates around 1.6 millions solutions per second on a regular PC, highly suitable for real-time vision tasks.
\item  Our method is simple to implement, which uses only about tens lines of C++ code (our source code is provided in appendix \ref{codes})--contrast to Ventura's machine-generated Groebner basis code of 16,000 lines \cite{Ventura05}.
\item  Our method is highly robust, capable of handling cases with extreme movements and dominating outliers.
\end{itemize} A quick comparison of our new method with several recent multi-camera relative pose methods is summarised in Table-\ref{table:compar}. \setlength{\tabcolsep}{4pt} 
{\small \begin{table}
\begin{center}
\caption{\small Time comparison of multi-camera relative pose solvers.}\vspace{-0.18in}
\label{table:compar}
\begin{tabular}{ |l|c|c||c|c|}
\hline 
Algorithm & $\sharp$ (Solutions) & time & $\sharp$ ({\small {RANSAC}} trials) & {\small RANSAC} time \\
\hline
{\bf 4pt-Ours  [this paper]}&{\bf 4} & {\bf 2.31$\upmu$s} & {\bf 143} &{\bf  0.16ms}\\
4pt-Sweeney \cite{Sweeney09}  & 6 & 12.78$\upmu$s & 143&1.83ms\\
4pt-Lee \cite{Lee10}  & 8 & 17.12$\upmu$s &143&2.45ms\\
6pt-Ventura \cite{Ventura05}  & 20 & 44.44$\upmu$s  &585&26.00ms\\
6pt-Stewenius \cite{Stewenius06} & 64 & 4.85ms &585& 2840ms\\
17pt-Li \cite{Li07}& 1 & 63.49$\upmu$s  &107214 &76650ms\\
\hline
\end{tabular}
\end{center}
\end{table}
\setlength{\tabcolsep}{1.4pt}}
\vspace{-0.4in}\subsection{Related work} 
The idea of deriving minimal case solvers for multi-camera relative pose is not new.  In \cite{Stewenius06}, Stewenius {\em et al.} proposed a minimal solution for the generalised epipolar equation using Groebner basis technique \cite{Cox12}.  Despite only minimally 6 points are used for estimating the 6-DOF motion, there are totally 64 solutions to choose from, rendering RANSAC inefficient.  Li {\em et al.} \cite{Li07} proposed several linear solvers for generalised camera relative pose, among which the most general case is the 17-points algorithm. However, the large number of feature points needed prevent it from being used with RANSAC.  Recently, Kneip and Li \cite{Kneip08} presented an iterative solution based on eigenvalue minimisation.  Although the objective function is in a closed form, its solution is sought via local iteration hence no converge guarantee.   Ventura et al.\cite{Ventura05} proposed to use first-order approximation of the camera rotation, by which the problem is reduced to solving a 20-th order polynomial.  Numerically, solving a 20-th degree polynomial is sensitive to noise. Using one known directional correspondence to simplify motion estimation is not uncommon.  Paper \cite{Naroditsky13}, proposed two efficient methods to solve the motion of a monocular camera. They give closed form solution to 4-order polynomial equation by utilising the fact that scale cannot be recovered in monocular vision. 

Mostly close related to our method is \cite{Lee10} in which the authors used IMU sensor to get two rotation readings and to solve the unknown yaw angles by solving an 8-th degree polynomial;  our experiments however shown that the method is  sensitive to image noise  and the reference direction noise.  In \cite{Sweeney09}, Sweeney et al. proposed the start-of-the-art method to solve the problem by artfully select the angle-axis representation of rotation matrix, and this results the maximum 6 real solutions. A drawback is that when all Pl\"{u}cker lines correspondences are from the same cameras in the first and second frames, numerical unstable problem happens when the rotation is small which is the usual case in real-world.  We do not compare methods using Ackerman vehicle motion model, as this model is restrictive in practice, and a post-relaxation is often needed (c.f. \cite{Generalized-Camera-Motion-Lee:CVPR-2013} \cite{Lee10}). 
%
%
\section{The basic idea for outliers removal: Conjugate Motion} 
In order to reliably estimate the vehicle's ego-motion from a non-overlapping multi-camera rig, a central task is to distinguish inliers (i.e. points matches on static backgrounds) from outliers (e.g. wrong matches or point matches on independently moving objects). In this paper, we propose to use the idea of {\em conjugate motion} to detect outliers. To explain this idea, let us look at the multi-camera system shown in the right-figure of fig-\ref{fig:config} for example, where a vehicle-mounted two-camera system observing two moving objects. In such a configuration, it is almost impossible for the two side-viewing cameras see the same moving object at the same time. The only possible shared `object' that can be seen by the two cameras simultaneously is the static background (e.g. road surface).  The relative motion between the vehicle and the static background is precisely the ego-motion that this paper aims to solver for.

When the outliers outnumber the inliers in the scene (e.g., imagine that an enormous truck is in the scene),  conventional algorithms can mistake outliers as inliers.  Let us use a numeric example to explain this: Suppose there are 100 feature matches found across two time-consecutive frames from each of the two cameras, and 80 matches are outliers from the moving objects, and 20 are inliers from the static background. In this case, a conventional monocular 8-point RANSAC algorithm may easily converge to outliers, as there are about 80\%  outliers but only 20\% inliers in the scene.  However, if one combines the information from both cameras' views, there is a simple way to detect outliers, regardless how high the outlier ratio is in each individual view. To be precise, let us use A to denote the vehicle's ego-motion {\em observed}  by the left camera (and represented in the left-camera's local frame), and use B to denote the ego-motion observed by the right camera (and represented in the right camera's local frame).  Then, it is well known that A and B must satisfy a conjugate motion relationship (also called as the ``hand-eye calibration" in robotics terminology), i.e. AX=XB, where X is the relative geometric calibration information between the two individual cameras, and is assumed known.  Therefore, by verifying whether or not A and B satisfy the conjugate relationship, we can tell whether or not they are possibly from the static background.  Here we assume that in general two {\em independently} moving objects do not satisfy the conjugate relationship, because they are {\em independent}.  In the above example,  only the 20\% background feature points can possibly satisfy the conjugate relationship.  This provides an efficient mechanism that allows relative pose algorithm to automatically focus its attention on the static background, thus significantly reduces the chance of getting distracted by those independently moving objects or outliers.
 
In our implementation,  we do not explicitly estimate A or B.  Instead, we directly estimate the relative pose by using feature correspondences from multiple individual cameras jointly using the Generalised Camera Model (GCM \cite{Using-many-as-one:CVPR-2003}).  This way, the conjugate relationship is automatically and implicitly enforced.
\section{A highly efficient minimal solver for GCM relative pose}
Our goal is to estimate the relative motion from two views captured by a calibrated multi-camera system in two successive time steps.  Six degree of freedoms are required to describe the 6-DOF Euclidean motion, such as the 6-point minimal solver in \cite{Stewenius06};  However it suffers from high computational complexity.

To simplify the computation, this paper makes two assumptions: (A) We note that for road vehicle application, often a common reference direction can be identified. This can be done by estimating the vertical vanishing point (if available), or by using the gravity direction provided by an IMU sensor, or assume the vehicle is driving on a locally flat plane (can be a slanted plane such as on a ramp); (B) We further assume the rotation is small; this is a reasonable assumption, because for real time driving application the video frame rate is necessarily high (e.g. greater than or equal to 10Hz); within a 0.1-second time interval a car cannot be turning by more than a few degrees.
\subsection{Generalised Camera Model}
The definition of generalised camera model can be found in \cite{Using-many-as-one:CVPR-2003}.  Given a calibrated GCM, we can obtain the normalised incoming ray direction corresponding to image point ${\bf{x}}_{i,j}$, where $i$ and $j$ are the camera and point index, respectively.  Denote $\bf u$ as the incoming ray and define a 6-vector pl\"{u}cker line  corresponding to the ray as ${\bf{l}}_{i,j} = \left[{{\bf{u}}}_{i,j}^T ,  \ \ \left({\bf{t}}_i\times {\bf{u}}_{i,j}\right) \right]$, where ${{\bf{t}}_i }$ is the intrinsic translation vector from camera $i$ to the GCM reference coordinate system. Then the generalised epipolar relationship is 
\begin{equation} \label{GECS}
{{\bf{l}}'^T_{i,j}}
\begin{bmatrix} {{{\left[ {\bf{ t}} \right]}_ \times }{\bf{R}}}, & {\bf{R}} \\ {\bf{R}},  & {\bf{0}} \end{bmatrix}
{{\bf{l}}_{i,j}} = 0,
\end{equation} where ${{\bf{l}}'^T_{i,j}}$ and ${{\bf{l}}_{i,j}}$ are matched pl\"{u}cker lines from the two frames 0 and 1, $\bf{t}$ and $\bf{ R}$ are the translation and rotation of the GCM reference coordinate system.
%
%
%
%
%
\subsection{A minimal 4-point algorithm using a directional correspondence}
To ease exposition, we assume an IMU sensor is available that provides one directional correspondence between two successive frames.  As explained above, using other methods (rather than IMU) is also possible. With this known vertical directional correspondence suppose we can obtain the roll and pitch angles. In this case, the only rotation angle that is left to be estimated is the yaw angle. Denote the corresponding rotation matrix between frame $0$ and frame $1$ as  $\left({\bf{R}}_y', {\bf{R}}_p', {\bf{R}}_r'\right) \leftrightarrow \left({\bf{R}}_y, {\bf{R}}_p, {\bf{R}}_r\right)$, where the subscript $\left( {y,p,r} \right)$ denote yaw, pitch and roll, and the corresponding pitch and roll rotation matrix rotate image rays direction to the reference direction.   At this stage, without loss of generality, we assume that the reference direction is the $z$ axis of the Earth coordinate system since both aligning the vertical vanishing point and directly exploiting IMU measurements will lead to the $z$ axis of the Earth coordinate system. Note that ${\bf{R}}_y'$ and ${\bf{R}}_y$ are inaccurate for exploiting IMU prior and unobservable for aligning the vertical vanishing points, which are the unknowns for solving. The relative rotation ${{\bf{R}}}_{ypr}$ can be written as
\begin{equation} \label{IMUwarp}
{{\bf R}}_{ypr} = {{\bf{R}}_r'^T}{{\bf{R}}_p'^T}{{\bf{R}}_y'^T}{{\bf{R}}_y}{{\bf{R}}_p}{{\bf{R}}_r}
\end{equation}
 
There is an implicit ambiguity in equation \eqref{IMUwarp}. Since ${{\bf{R}}_y'^T}{{\bf{R}}_y}$ can be merged into one yaw rotation matrix which is denoted as ${{ \bf{\Delta R}}_y}$, only the relative yaw rotation matrix can be solved. 

Denote ${{\bf{R}}_p}{{\bf{R}}_r}$ and ${{\bf{R}}_p'}{{\bf{R}}_r'}$ as ${{\bf{R}}_{pr}}$ and ${{\bf{R}}_{pr}'}$, respectively and substituting equation \eqref{IMUwarp} into equation \eqref{GECS} gives:
{\small \begin{equation} \label{IMUwarp3}
{\left( {\left[ {\begin{array}{*{20}{c}}
{{{{\bf{R}}}_{pr}'},}&{\bf{0}}\\
{{\bf{0}},}&{{{{\bf{R}}}_{pr}'}}
\end{array}} \right]{\bf{l'}}} \right)^T}\left[ {\begin{array}{*{20}{c}}
{{{\left[ {{\bf{\tilde t}}} \right]}_ \times } {{\bf{\Delta R}}_y},}&{ {{\bf{\Delta R}}_y}}\\
{ {{\bf{\Delta R}}_y},}&{\bf{0}}
\end{array}} \right]\left[ {\begin{array}{*{20}{c}}
{{{\bf{R}}_{pr}},}&{\bf{0}}\\
{{\bf{0}},}&{{{\bf{R}}_{pr}}}
\end{array}} \right]{\bf{l}} = 0,
\end{equation}}
where ${\bf{\tilde t}} = {{{\bf{R}}}_{pr}'}{\bf{t}}$ and the subscript $i,j$ is dropped for simplicity.

In equation \eqref{IMUwarp3}, the unknowns are $\bf \tilde t$ and ${\bf \Delta R}_y$. Utilizing the small relative rotation assumption, we apply the first-order approximation to ${\bf \Delta R}_y$, parameterizing it by a 3-vector ${{{\bf{ r}}}_y} = {\left[ {\begin{array}{*{20}{c}}
0,&0,&{{r_{y}}}
\end{array}} \right]^T}$:

\begin{equation}\label{R}
{{{\bf{\Delta R}}}_y} \approx {\bf{I}} + {\left[ {{{{\bf{ r}}}_y}} \right]_ \times }
\end{equation}
We parameterize ${{\bf{ \tilde t}}}$ as
\begin{equation}\label{t}
{\bf{\tilde t}} = {\left[ {\begin{array}{*{20}{c}}
{{{ t}_x}},&{{{ t}_y}},&{{{ t}_z}}
\end{array}} \right]^T}
\end{equation}

Substituting equation \eqref{R}, \eqref{t} into equation \eqref{IMUwarp3} gives 
\begin{equation} \label{ploy}
{a_1} + {a_2}{{ t}_x} + {a_3}{{ t}_y} + {a_4}{{ t}_z} + {a_5}{r_{y}} + {a_6}{{ t}_x}{r_{y}} + {a_7}{{ t}_y}{r_{y}} + {a_8}{{ t}_z}{r_{y}} = 0
\end{equation}
where $a_1$ to $a_8$ are the coefficients formed with the pl\"{u}cker line correspondence ${\bf{ l'}} \leftrightarrow {\bf{ l}}$, parameterized by a 6-vector ${\bf{  l'}} = {\left[ {\begin{array}{*{20}{c}}
{{{  l}_x'}},&{{{  l}_y'}},&{{{  l}_z'}},&{{{  l}_u'}},&{{{  l}_v'}},&{{{  l}_w'}}
\end{array}} \right]^T}$ and  ${\bf{  l}} = {\left[ {\begin{array}{*{20}{c}}
{{{  l}_x}},&{{{  l}_y}},&{{{  l}_z}},&{{{  l}_u}},&{{{  l}_v}},&{{{  l}_w}}
\end{array}} \right]^T}$, respectively.
%
%
In order to solve for the four unknowns ${{ t}_x},{{t}_y},{{ t}_z}$, and ${r_{y}}$, one requires minimal four pl\"{u}cker line correspondences. This gives rise to a system of four polynomials with the other three polynomials in similar form as equation \eqref{ploy} with the coefficients denoted by $b_1$ to $b_8$, $c_1$ to $c_8$ and $d_1$ to $d_8$.  After stacking all four correspondences and separating $r_y$ from ${{ t}_x},{{t}_y},{{ t}_z}$, we arrive at an equation system
{\small 
\begin{equation} \label{rt}
\underbrace {\left( {\begin{array}{*{20}{c}}
{{a_2} + {a_6}{\mkern 1mu} {r_y}}&{{a_3} + {a_7}{\mkern 1mu} {r_y}}&{{a_4} + {a_8}{\mkern 1mu} {r_y}}&{{a_1} + {a_5}{\mkern 1mu} {r_y}}\\
{{b_2} + {b_6}{\mkern 1mu} {r_y}}&{{b_3} + {b_7}{\mkern 1mu} {r_y}}&{{b_4} + {b_8}{\mkern 1mu} {r_y}}&{{b_1} + {b_5}{\mkern 1mu} {r_y}}\\
{{c_2} + {c_6}{\mkern 1mu} {r_y}}&{{c_3} + {c_7}{\mkern 1mu} {r_y}}&{{c_4} + {c_8}{\mkern 1mu} {r_y}}&{{c_1} + {c_5}{\mkern 1mu} {r_y}}\\
{{d_2} + {d_6}{\mkern 1mu} {r_y}}&{{d_3} + {d_7}{\mkern 1mu} {r_y}}&{{d_4} + {d_8}{\mkern 1mu} {r_y}}&{{d_1} + {d_5}{\mkern 1mu} {r_y}}
\end{array}} \right)}_{{\bf{M}}\left( {{r_{y}}} \right)}\left[ {\begin{array}{*{20}{c}}
{{{  t}_x}}\\
{{{  t}_y}}\\
{{{  t}_z}}\\
1
\end{array}} \right] = {\bf{0}}
\end{equation}} 
Since ${\bf{M}}\left( {{r_{y}}} \right)$ is a square matrix and it has a non-trivial solution when the determinant of ${\bf{M}}\left( {{r_{y}}} \right)$ is zero, consequently: 
{\small \begin{equation} \label{polyno}
Ar_{y}^4 + Br_{y}^3 + Cr_{y}^2 + Dr_{y}^{} + E = 0,
\end{equation}} where $A,B,C,D$ and $E$ are formed from the coefficients $a,b,c$ and $d$ of the system of polynomials. Because the univariate polynomial \eqref{polyno} is 4-th order,  it admits closed form solutions with maximum 4 real roots (its analytic solutions are given in appendix \ref{closedSOl}).  Once rotation is found, to solve for translation we simply use QR decomposition of the above equation system. 
%
%
%
%

\subsection{Preemptive RANSAC sampling}
RANSAC (Random Sample Consensus) is a simple, yet powerful tool for solving a variety of robust estimation problems in computer vision. To achieve robust relative pose in highly dynamic environment, we plug our new 4-point algorithm into a RANSAC framework. However, in order to exploit the conjugate relationship for faster outlier-pruning, we made a small but necessary modification to the sampling scheme of the conventional RANSAC. We note that the conventional RANSAC algorithm generates hypotheses {\em uniformly} in the sense that it only samples the input dataset uniformly, without using any prior information (or special structure) of the problem instance. In practice, this may be an overly pessimistic process, since a priori information is often available and can be used to generate better hypotheses. In our case, we wish to {\em preemptively} discard those unpromising motion hypotheses as early as possible. We propose a simple modification to the RANSAC sample process, i.e., we require that among every minimal set of 4 points at least two points are taken from two different cameras with non-overlapping FOVs.  By this, we ensure that the candidate motion solved by the 4-point set are compatible (i.e. satisfying the {\em Conjugate Motion} relationship). This way, the sampling procedure is free from getting points all from the same camera view, which would be problematic if the outliers outweigh inliers in that view. 

Our sampling scheme is simply and intuitive. We are fully aware of that, in the vast body of RANSAC literature there must be a similar idea published (e.g.,\cite{RANSACREF}). Despite this, in this paper rather than looking into the literatures we take a pragmatic solution and to solve our current problem as quickly as we can, and our present preemptive sampling scheme is such a one which works surprisingly well in practice; it quickly eliminates up to 80\%--90\% wrong matches on large moving objects, which otherwise would be detected as inliers by conventional methods. Overall, our ``4-point+RANSAC" method is shown to be efficient in removing independent motions in all our experiments. Note also, despite having the same name, our preemptive RANSAC sampling scheme is different from Nister's ``preemptive RANSAC" \cite{NisterRANSAC} as they have different motivations.
\vspace{-0.2in}\subsection{Discussions}\paragraph{Scale recovery.} Although it is well known that a multi-camera rig in generic motions is able to estimate a full 6-DOF motion with absolute scale \cite{Lee10},  we find our method falls short in scale recovery.  This is however understandable for two reasons: 1) our method only uses feature correspondences from the same individual camera, and 2) our small rotation assumption suggests the motion is very close to pure translation, which is known to be a degenerate case for scale recovery.  In order to recover the absolute scale, one needs to use additional information (e.g., cross-camera correspondences) whenever available.  \vspace{-0.12in}\paragraph{Pure rotation or planar scene.}Our method can handle successfully pure rotation case,  thank to the decoupling of rotation from translation as shown in  Eq. \eqref{rt}. In our simulated experiments we did not observe any numerical instability issue when ground-truth translation is close to zero.  Following a similar test, we also confirm our method is robust to pure planar scene structure.  We have evaluated extensively on cases where all 4 points are co-planar or co-linear, and there were no degeneracies observed. 
 \vspace{-0.12in}\paragraph{Ground plane motion.} Our method does not assume the vehicle driving on the ground plane, it works for most situations including on a road ramp or uphill, i.e. cases which would defy methods based on Ackerman motion model (e.g.\cite{David1ptRANSAC}). 
\section{Experiments}
We conducted experiments on both simulated data and real images to validate the proposed method. In our simulations, we test the robustness of our method w.r.t pixel noise level, errors in the directional reference, and magnitude of rotation angles. We simulated a 2-camera rig system with 1$m$ baseline and non-overlapping field of views. We create random 3D points in space and generate matches. The motion of the camera rig is randomly chosen under a condition that the rotation angle is below 5 degrees. We purposely set camera's field of view and relative distances to scene to mimic that in KITTI autonomous driving datasets \cite{KITTI-Dataset:CVPR-2012}. We only use feature correspondences from the same camera in two consecutive frames.  We implemented our algorithm in Matlab (with C/C++ mex interface), and tested it on a regular PC of 3.4Ghz.  Our method is extremely simple to code,with only tens of lines of source code (we provide the code in supple. material).  

\subsection{Robustness to Pixel Noise}
\begin{figure}
\centering
\includegraphics[width=0.37\textwidth]{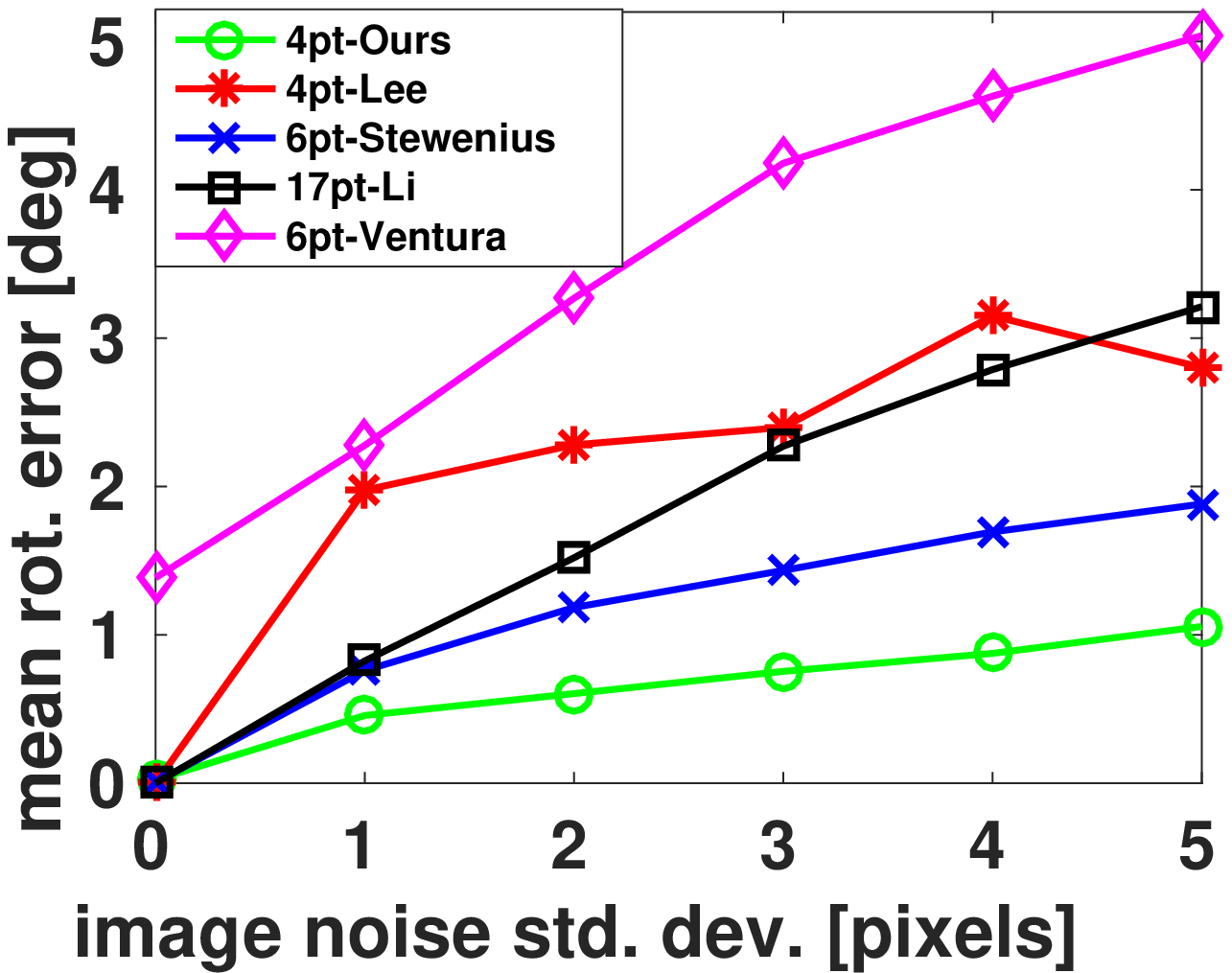}~~~~~~~~
\includegraphics[width=0.37\textwidth]{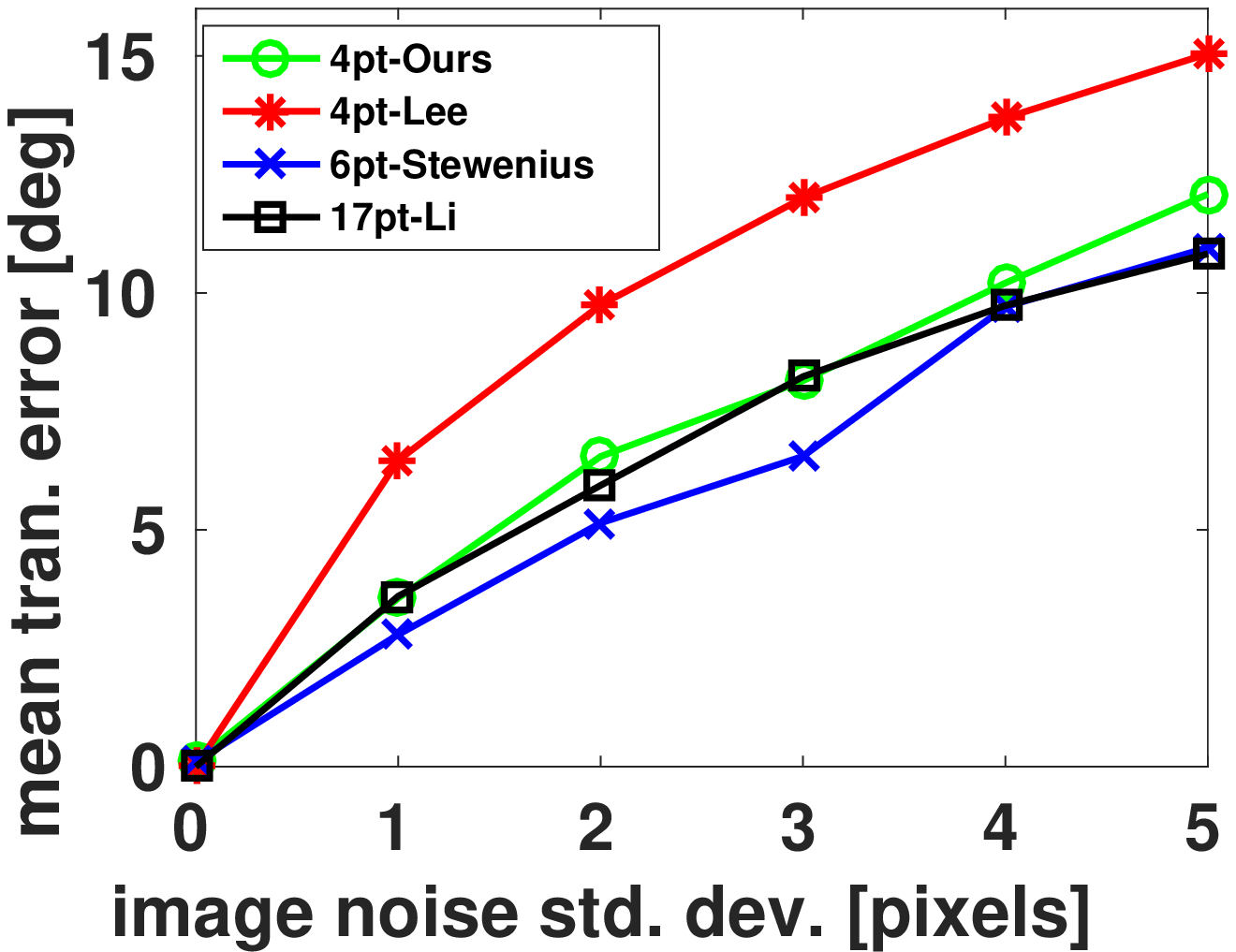}
\caption{{\small Method comparison under different levels of image pixel noise.}}
\label{fig:pixelnoise}
\end{figure}

To measure the accuracy and robustness of our method under different image noise level, we add Gaussian noise to feature coordinates, with standard deviation ranging from 0 to 5 pixels. We set the ground-truth rotation angle at about $1$ degree per frame to reflect small rotation condition. It is also a realistic value for road vehicle driving at regular speed, as we have observed in the KITTI datasets \cite{KITTI-Dataset:CVPR-2012}.  The experiment results obtained from 1,000 random tests are given in fig-\ref{fig:pixelnoise}. As can be seen, the rotation estimate accuracy (from only 4 points) by our method is the highest, though our translation direction estimation is slightly inferior to 6pt-Stewenius 6pt \cite{Stewenius06} and 17pt-Li \cite{Li07} linear methods. The average running time for our method is about 2.3$\upmu s$, while the 6pt algorithms takes 4.8$ms$, i.e. 2,000 times slower.
\subsection{Robustness to noise in the reference direction}
\begin{figure}
\centering
\includegraphics[width=0.4\textwidth]{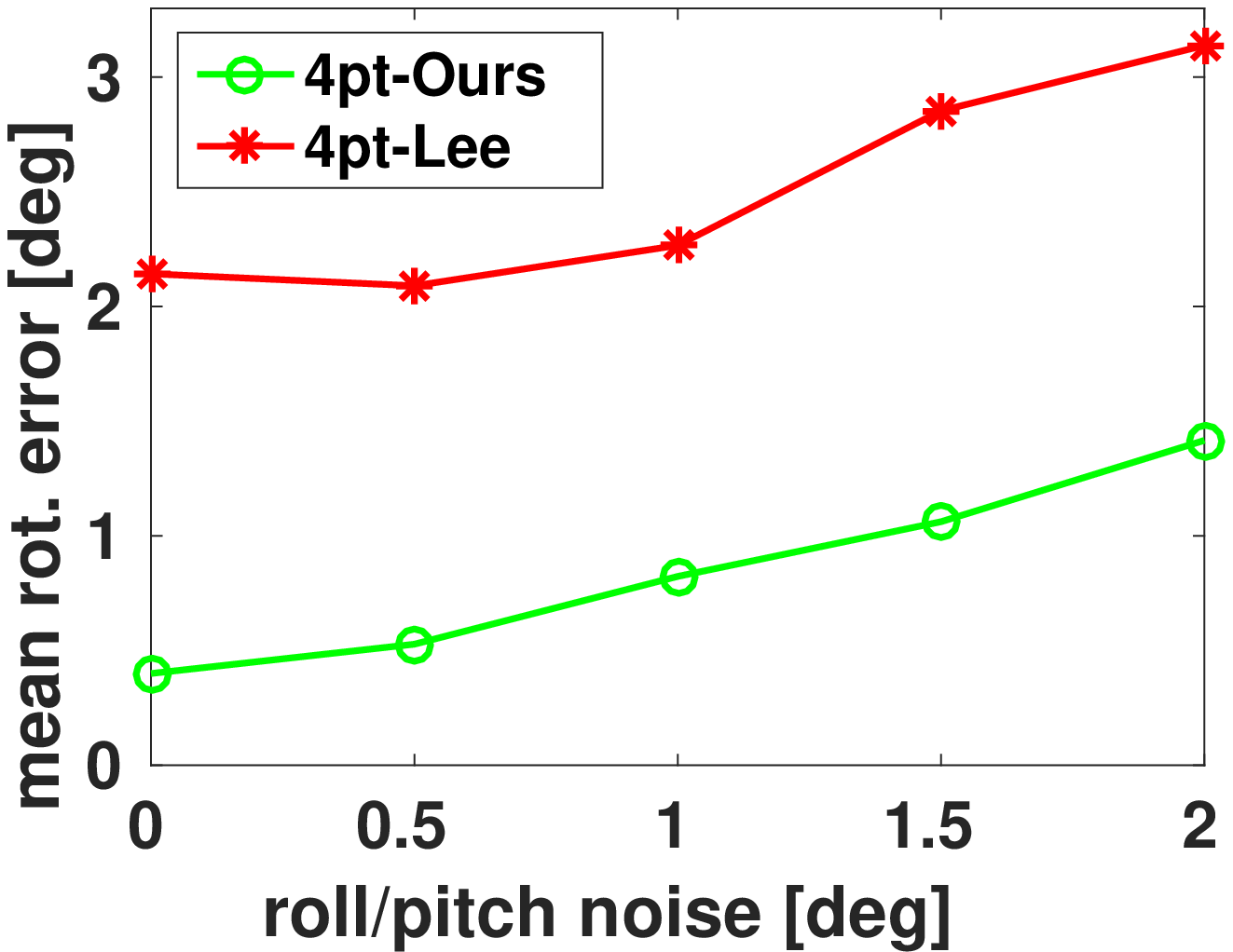}
\includegraphics[width=0.4\textwidth]{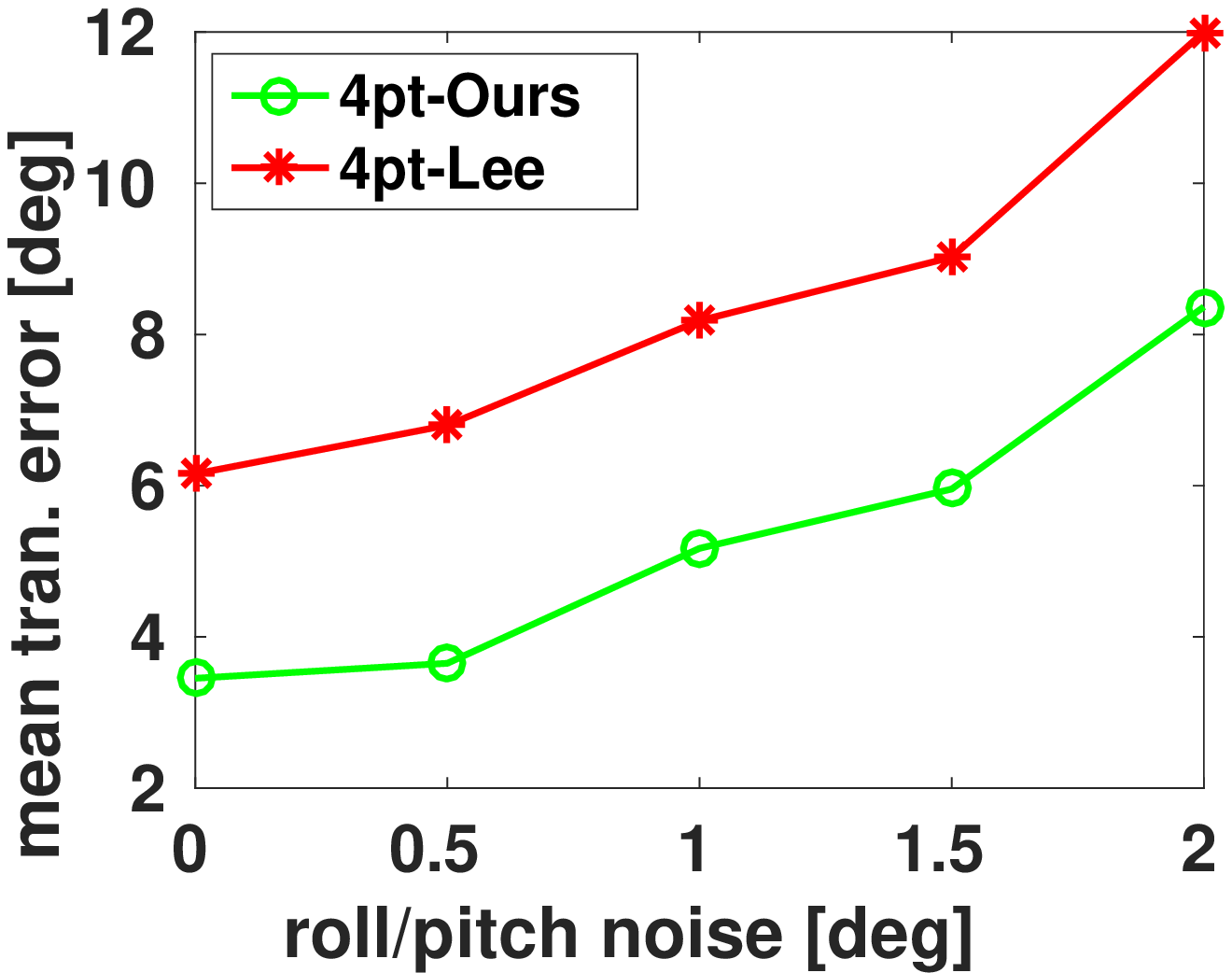}
\caption{Robustness of our method w.r.t. errors in the directional reference.}
\label{fig:directionnoise}
\end{figure}

Since our method rely on a known reference direction, one would like to investigate the robustness of it w.r.t to errors in the reference direction.  We add Gaussian noise to the roll and pitch angles inferred from the reference direction. We chose noise level comparable to today's consumer-grade IMU found on typical smartphones (e.g. the nominal noise of Xsens MTi IMU sensor in roll/pitch is about 0.3 degrees\cite{MTXSENS}).  The test results are given in fig-\ref{fig:directionnoise}, and we only compare our method with 4pt-Lee \cite{Lee10} method as it also relies on a reference direction. As can be seen, our method outperforms at all levels of roll/pitch noise.
\subsection{Robustness to the magnitude of ground-truth rotations}
\begin{figure}
\centering
\includegraphics[width=0.4\textwidth]{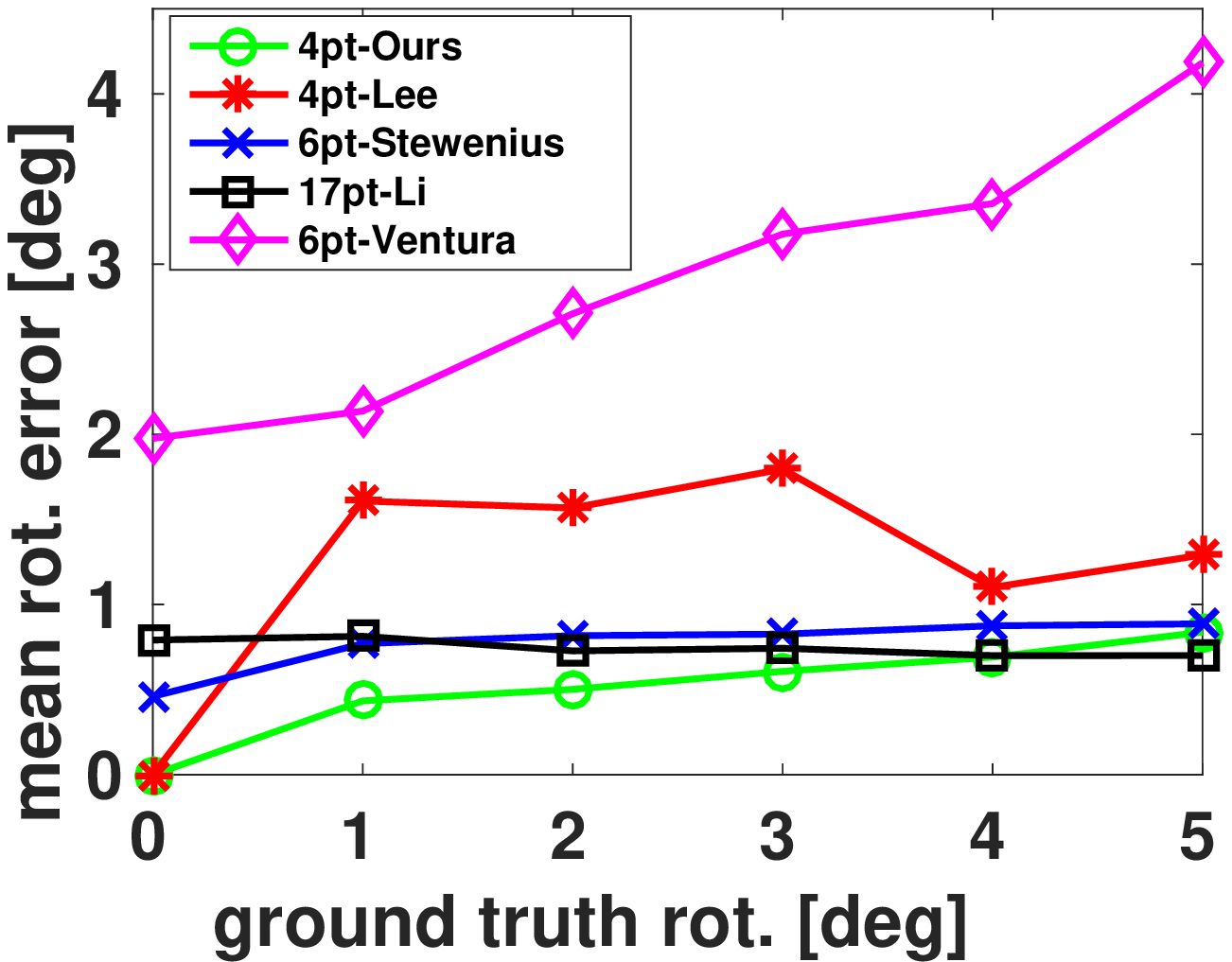}
\includegraphics[width=0.4\textwidth]{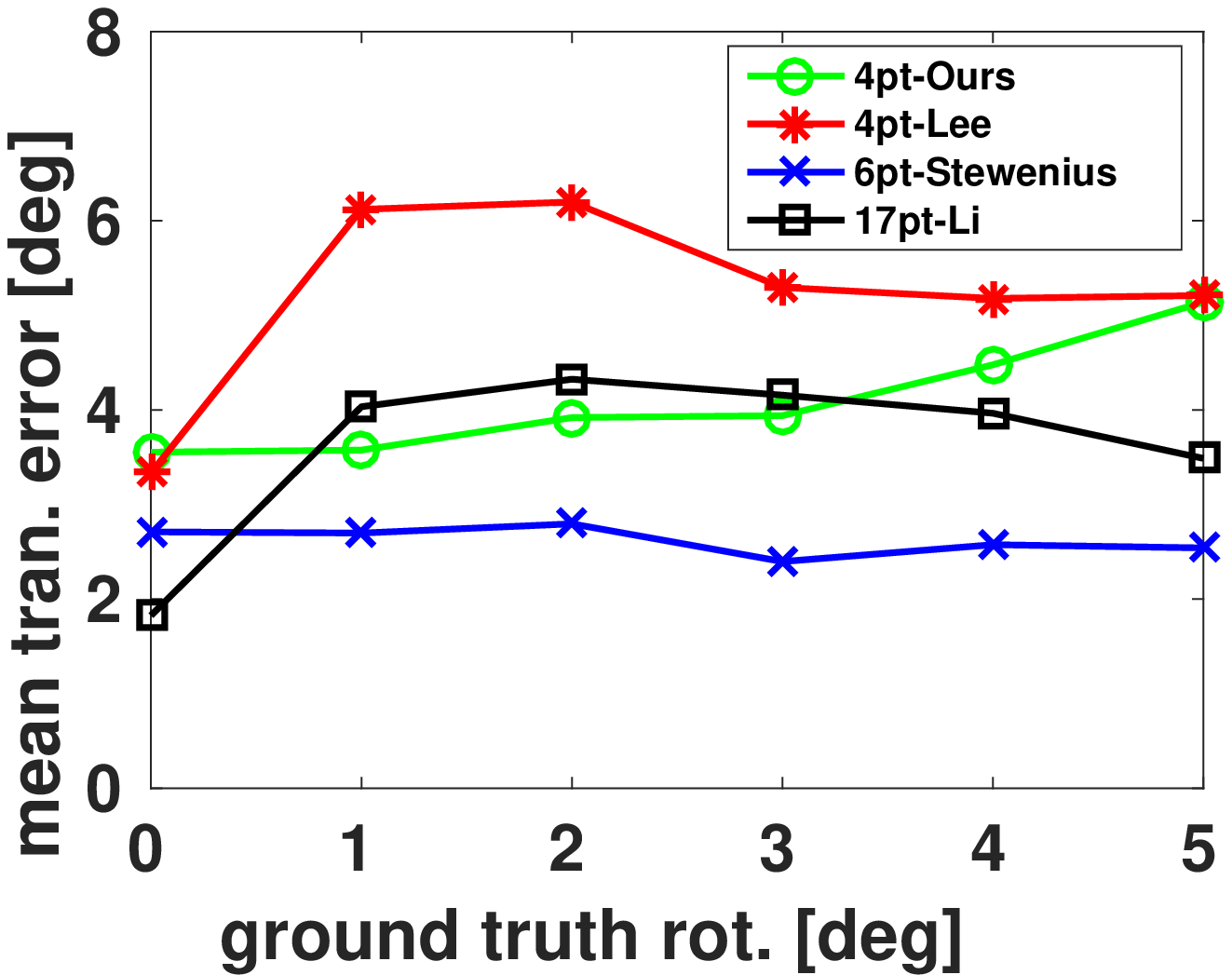}
\caption{\small Method Comparison under different rotation angles (to validate the small-rotation assumption).}
\label{fig:nonoise}
\end{figure}

The goal of this test is to evaluate the efficacy of our small rotation assumption, and to check 
the numerical stability of our method when the actual rotation angle is big. In this test, we also applied noise to image pixels, but no noise was added to the reference direction. We increase the ground-truth rotation angle and compare the results with several state-of-the-art methods: i.e., the 6pt-Stewenius \cite{Stewenius06}, 17-point-Li \cite{Li07}, 6pt-Ventura \cite{Ventura05} and the 4pt-Lee in \cite{Lee10}. We vary rotation angles in $[0\ 10]$ degrees, and generate 1,000 random instances for averaging. Fig-\ref{fig:nonoise} shows a summary comparison of the above methods in terms of the average rotation and translation direction errors. The rotation errors are computed by $\epsilon=\arccos((\mathbf{trace}({\bf{R}}^{T}_{gt}{\bf{R}}_{est})-1)/{2}),$ where ${\bf{R}}^{T}_{gt}$ is the ground truth rotation and ${\bf{R}}_{est}$ the estimated rotation, respectively.
In the rotation estimation, our method outperform all methods in small rotation angles, but the accuracy degrades gradually with increasing angles for the first-order approximation. Both 6pt-Stewenius \cite{Stewenius06} and 17pt-Li \cite{Li07} gives stable estimations. Although our method and 6pt-Venture \cite{Ventura05} all use the small rotation assumption, our method degrades more gracefully. The reason we think is due to that 6pt-Venture \cite{Ventura05} needs to solve a unstable 20-degree polynomial, while ours is only 4. The same reason may apply to  4pt-Lee \cite{Lee10} which involves an 8-th order polynomial. Paper \cite{Naroditsky13} also used 4-th order polynomial, but it is only applicable to monocular camera. Our results for translation (direction) estimation is slightly worse than 6pt-Stewenius, but is comparable with or better than the others.
\subsection{RANSAC Estimation}
In this section, we did two sets of experiments to validate our method's robustness in high ratios of outliers. In all the following tests, Gaussian noise of stdv=1 is added to pixels, and 0.5 deg noise is added to roll/pitch angles. The first experiment is to set the inlier ratio $w$ to $0.5$, the RANSAC confidence level $p$ is set at $0.9999$; the number of iterations is given by $k=\frac{\ln \left(1-p\right) }{\ln \left(1-w^{n} \right) }$ of each method to check the performances, where $n$ is the minimal number of correspondences needed.

\begin{figure}
\centering
\includegraphics[width=0.45\textwidth]{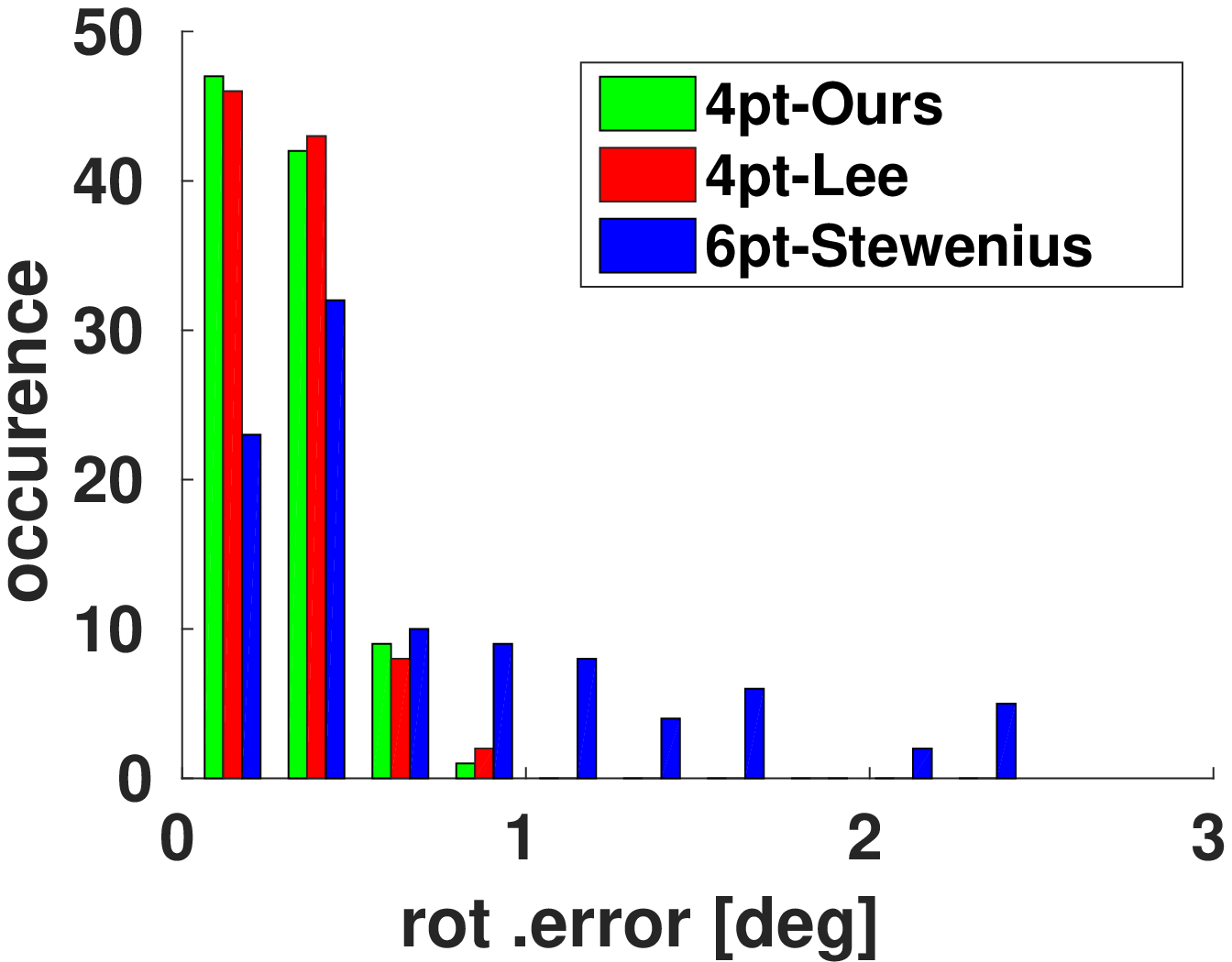}
\includegraphics[width=0.45\textwidth]{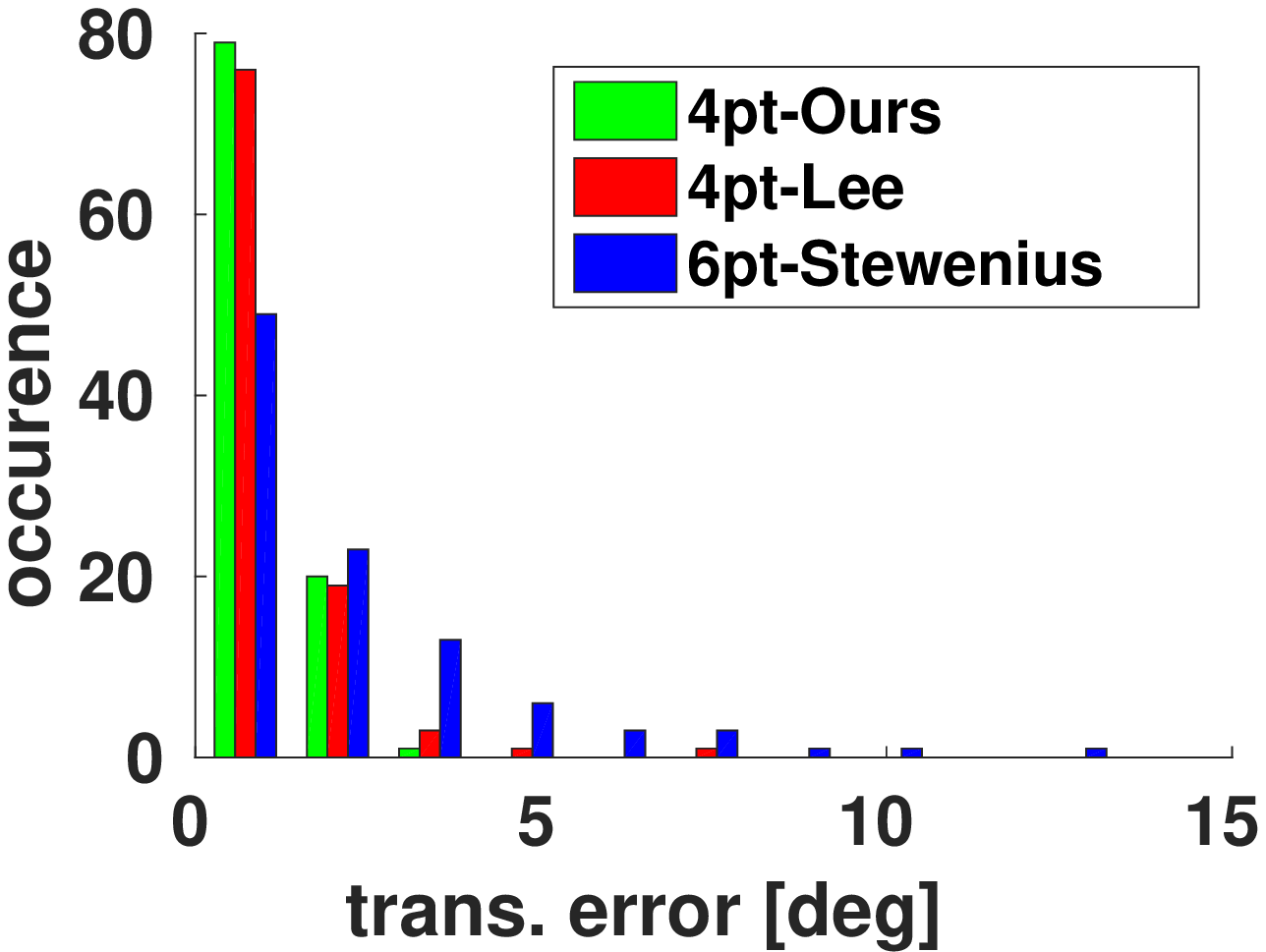}
\caption{\small RANSAC experiment result for fixed outlier ratio at 50\%, and fixed inlier-confidence level at $99.99\%$. Average result of 100 independent runs. Our method clearly outperforms all competing methods in the figure.}\label{fig:ransac_adaptive}
\end{figure}

The results are given in figure \ref{fig:ransac_adaptive}. It can be seen that the rotation and translation estimation of the 4-points methods are similar, while our method is slightly better. The 6-pt Stewenius \cite{Stewenius06} leads to higher rotation and translation error.

\begin{figure}
\centering
\includegraphics[width=0.32\textwidth]{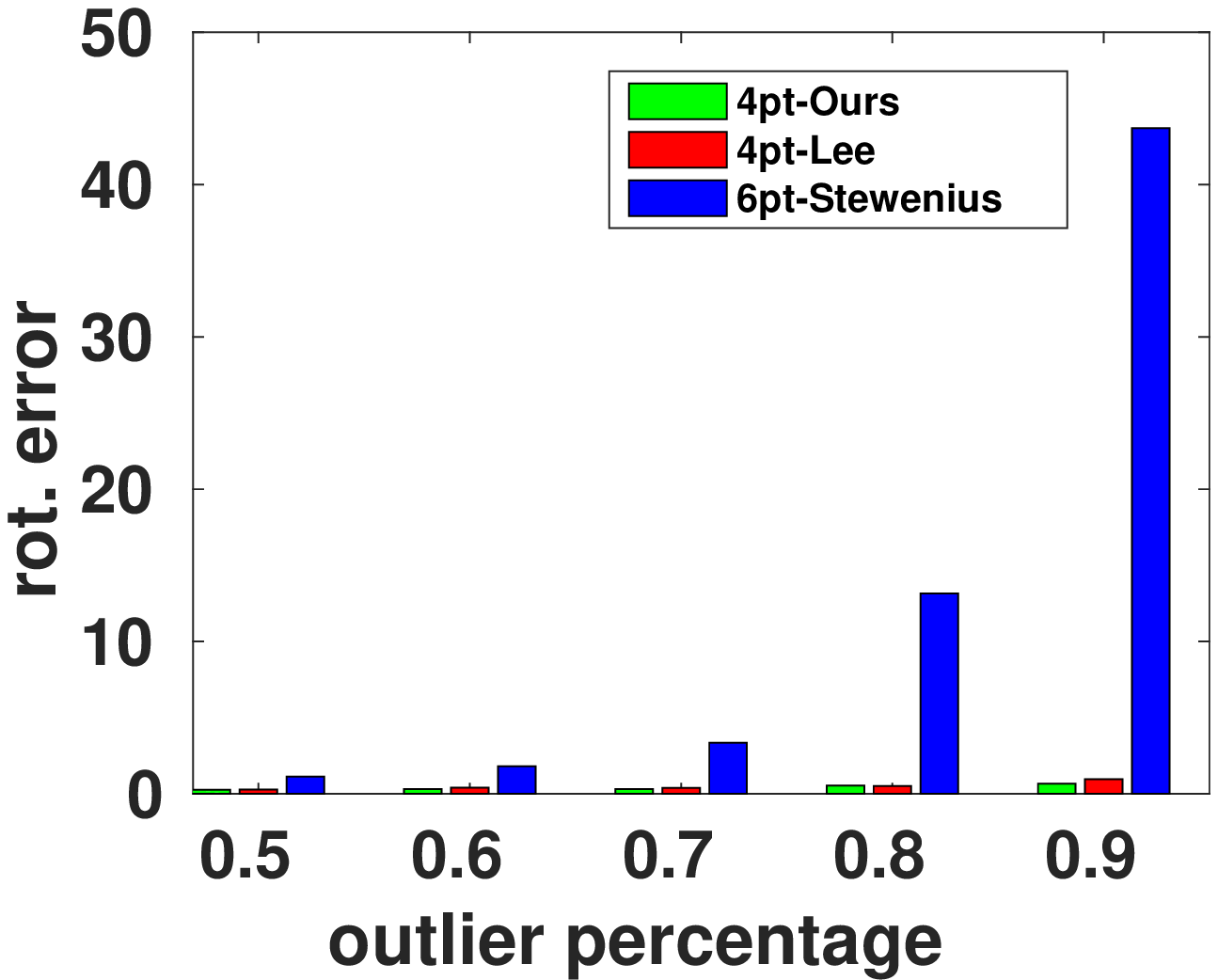}
\includegraphics[width=0.32\textwidth]{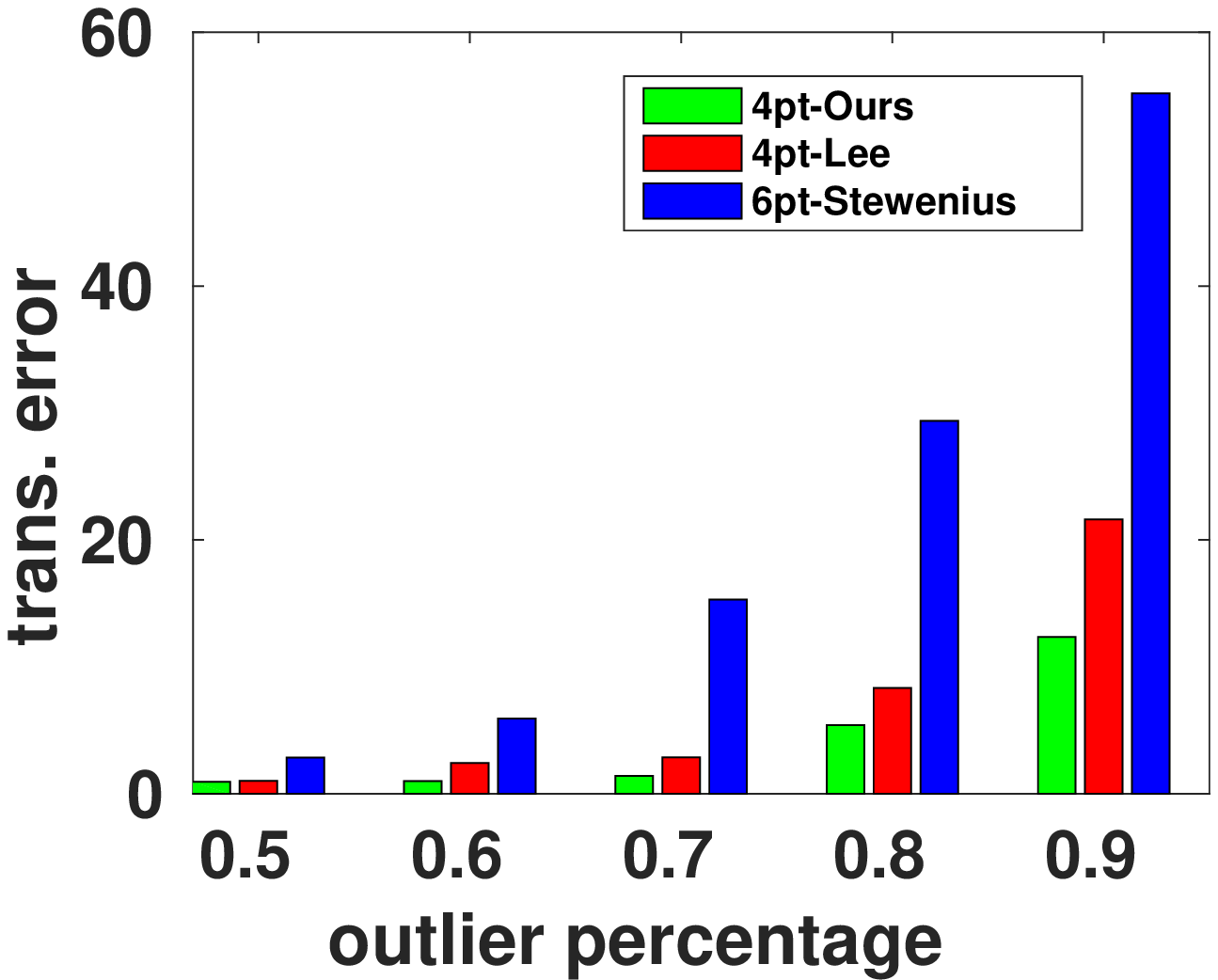}
\includegraphics[width=0.32\textwidth]{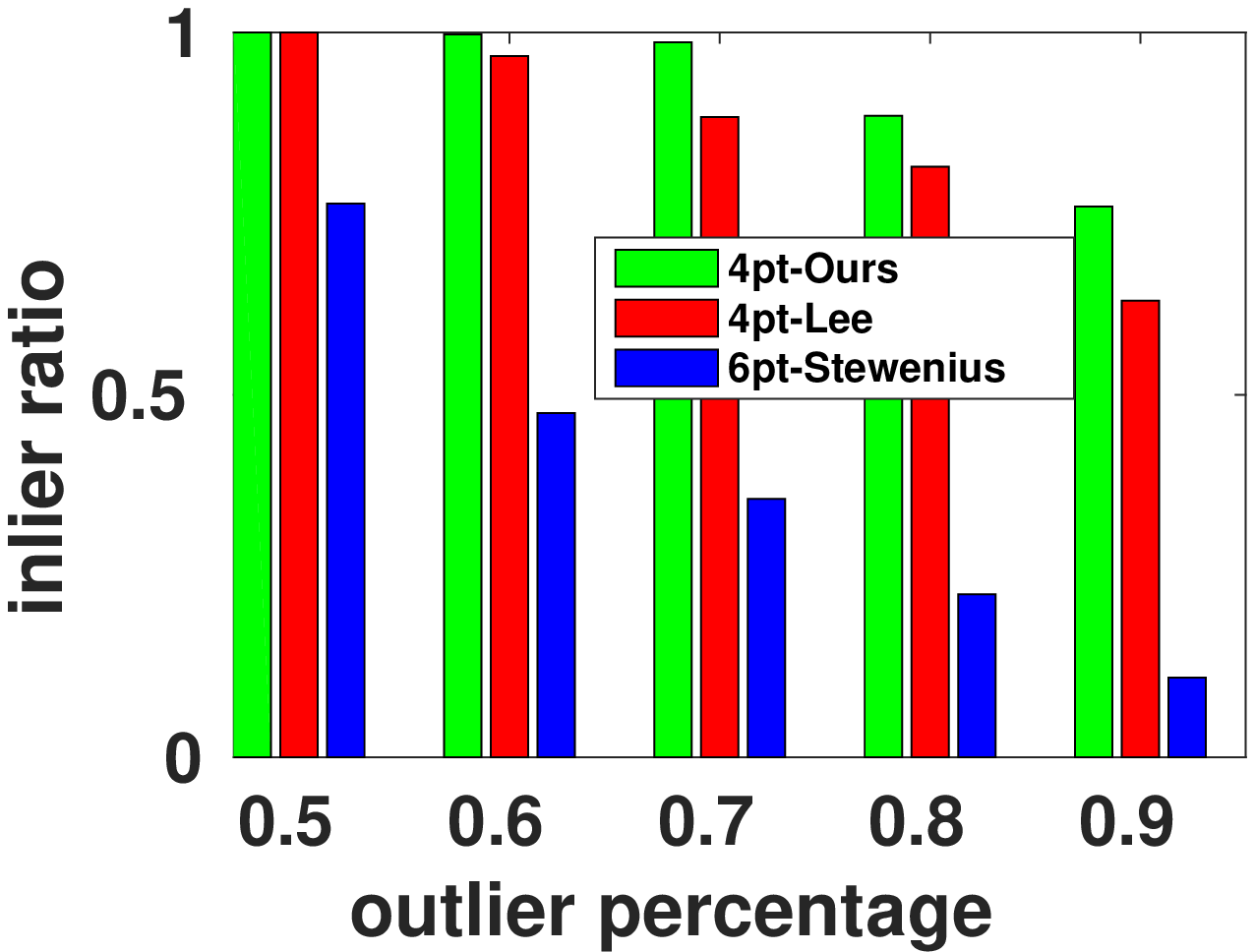}
\caption{\small RANSAC experiments with fixed (and same) iteration numbers at 500.  Our method achieves both lower error-rate and higher inlier success-rate.} 
\label{fig:ransac_outliers}
\end{figure}

The second experiment we did is to test the RANSAC with fixed and same iteration number (e.g. 500 times), under varying and high outlier ratios. We vary the outlier ratio from $50\%$ to $90\%$ and use a fixed iteration number of $500$ to compare the performances. Fig-\ref{fig:ransac_outliers} shows our method successfully selects all inliers for interval $[50\%,70\%]$ while others all had some missed out. There is a noticeable performance decrease from outlier percentage at $80\%$, the reason may be that the iteration we used is too small at this outlier ratio level.


\subsection{Experiments on real images}
\subsubsection{ Lab-based experiment.}  
\begin{figure}[h!]
\centering
\subfigure[configuration]{%
\includegraphics[width=0.37\textwidth]{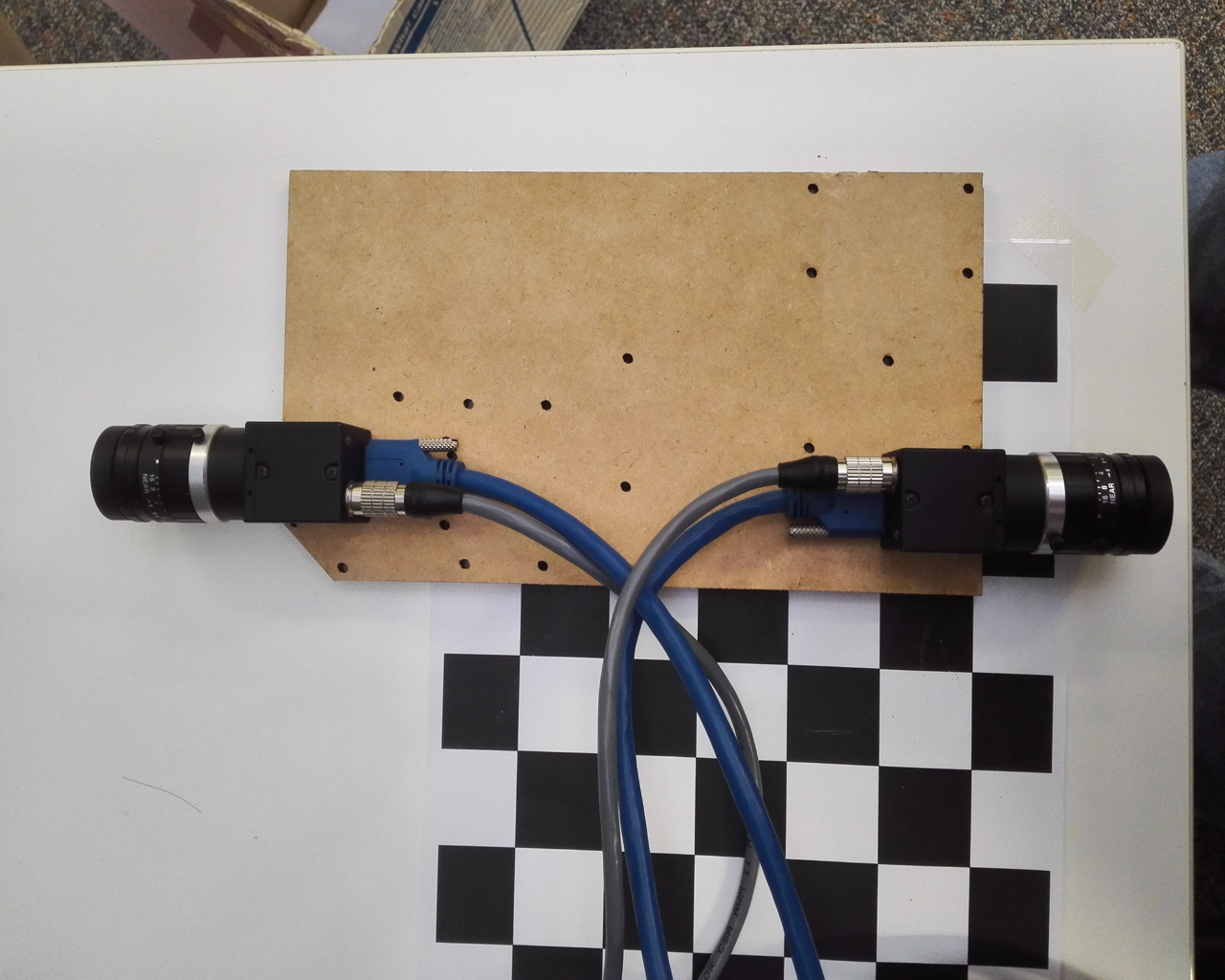}~~
\label{sys_configuration}}
\subfigure[a sample image]{%
\includegraphics[width=0.37\textwidth]{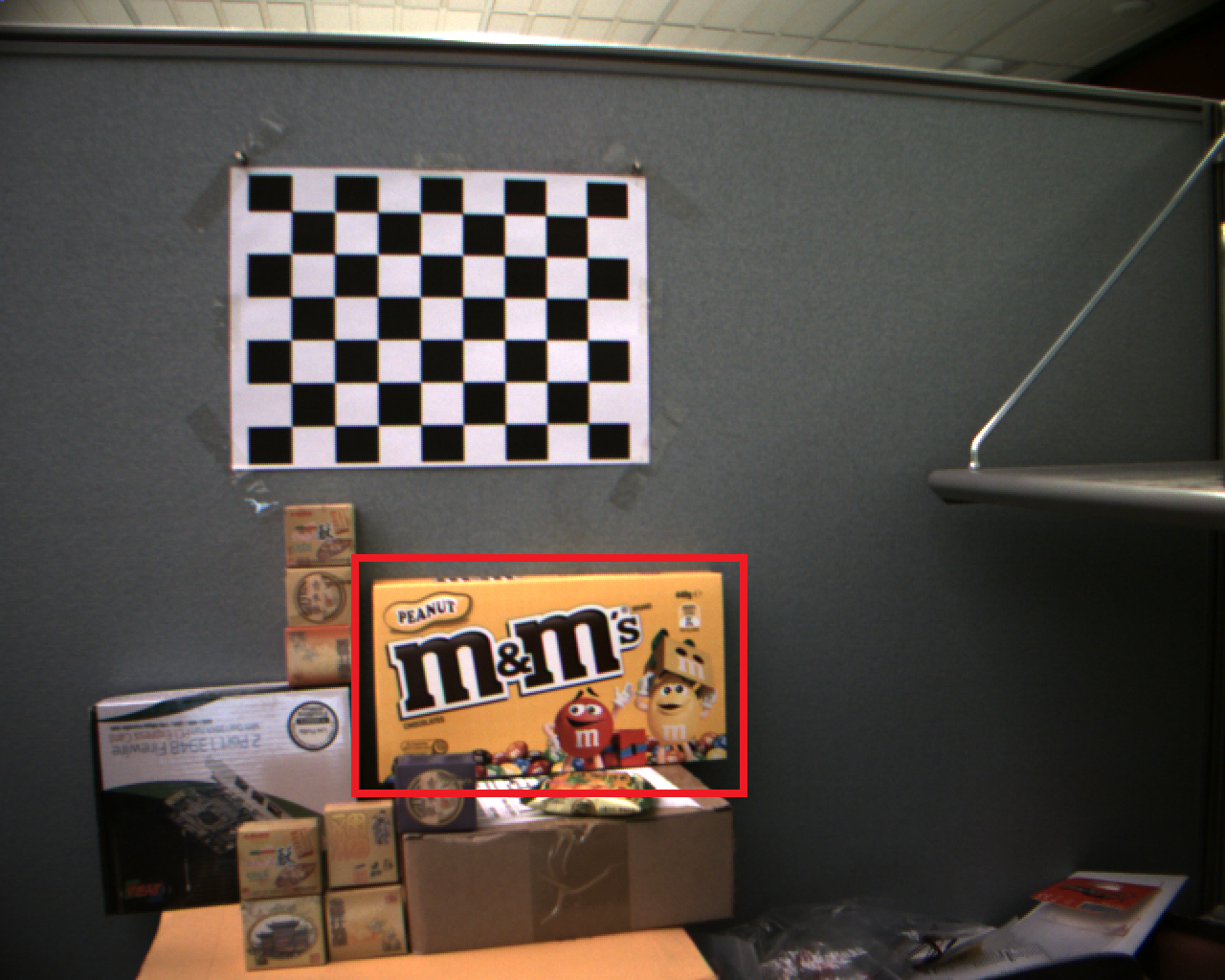}
\label{sys_sampleimage}}\\

\subfigure[our method (image 0)]{%
\includegraphics[width=0.38\textwidth]{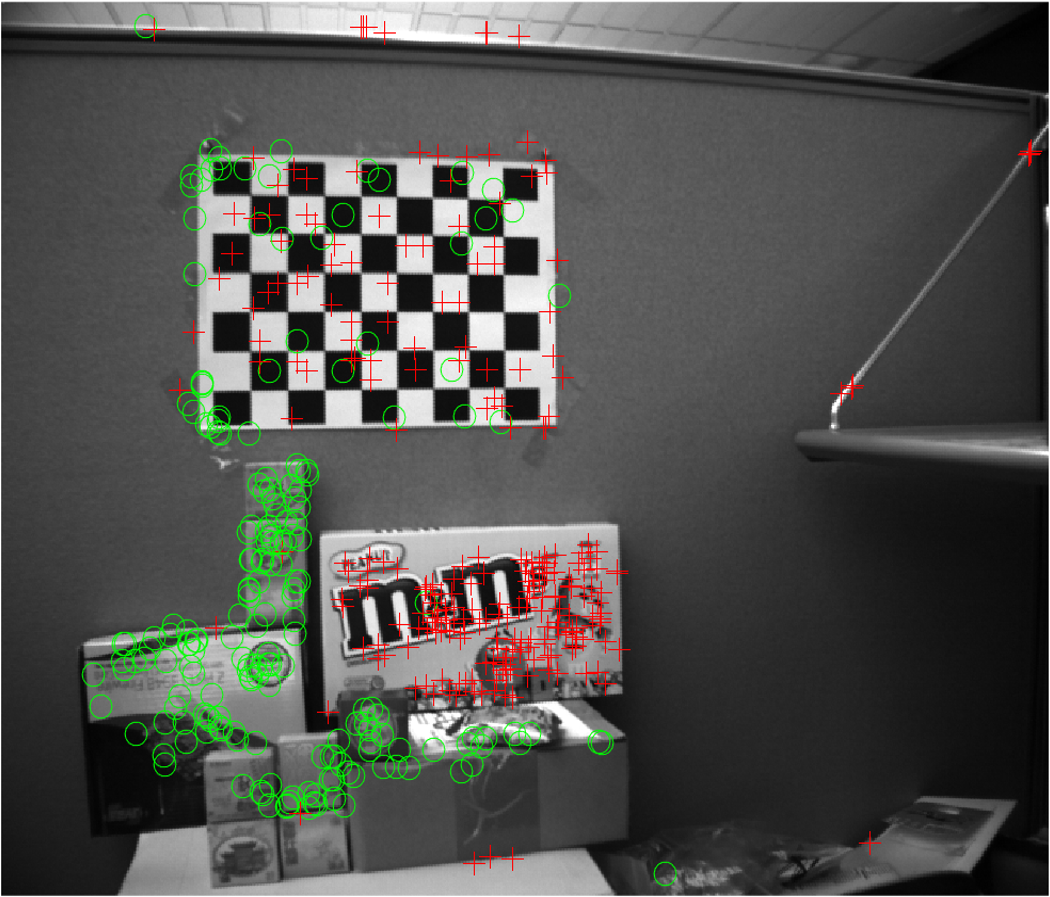}
\label{sys_our0}}
\subfigure[our method (image 1)]{%
\includegraphics[width=0.38\textwidth]{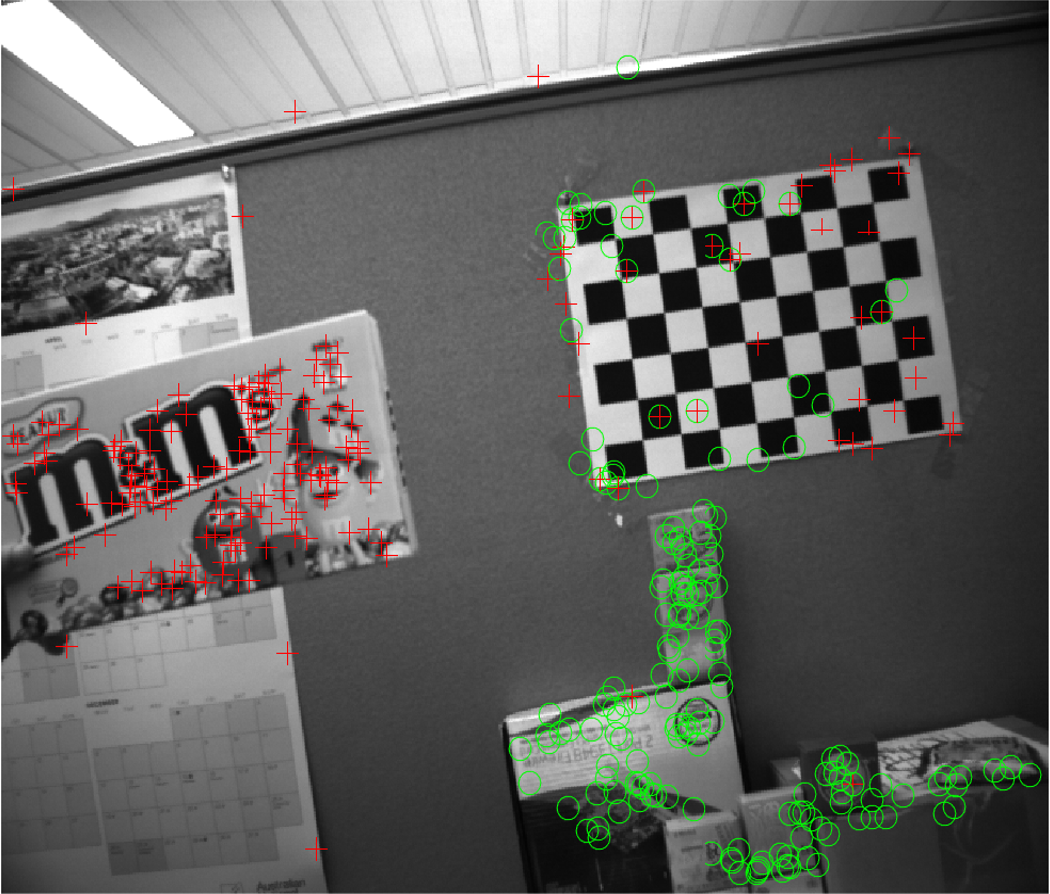}
\label{sys_our1}}
\subfigure[8-point method (image 0)]{%
\includegraphics[width=0.38\textwidth]{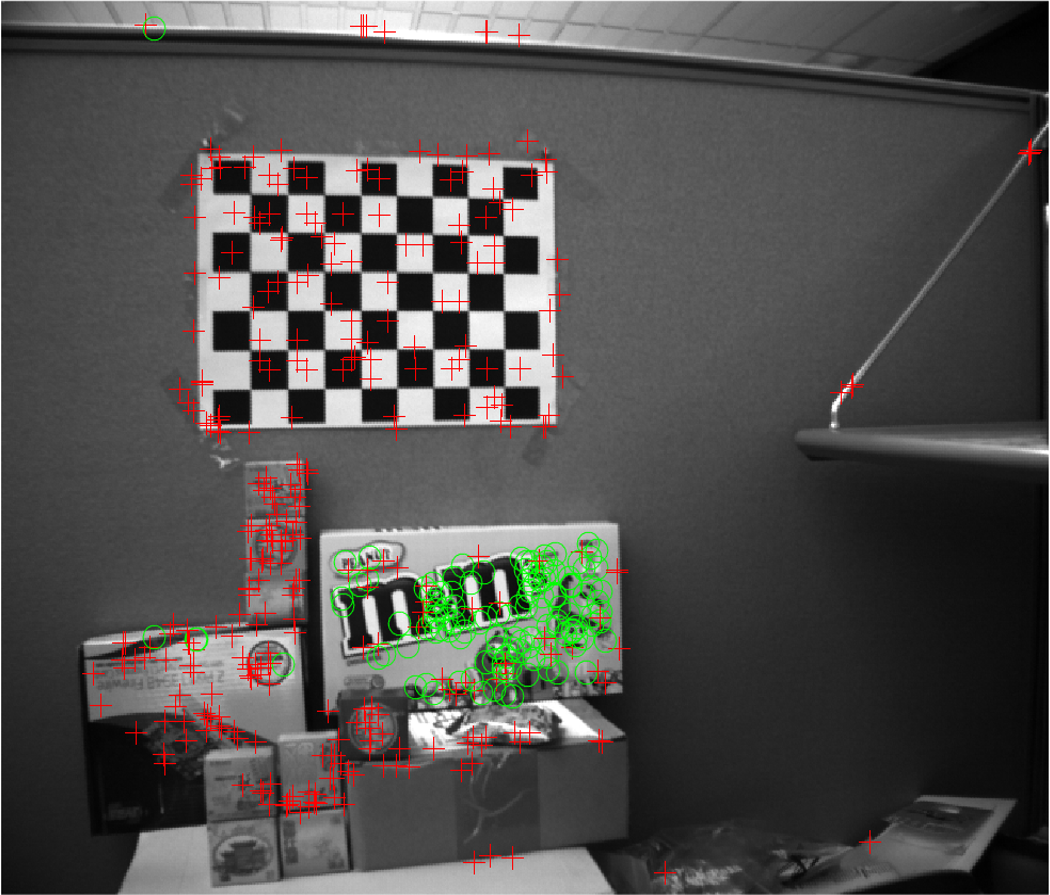}
\label{sys_8pt0}}
\subfigure[8-point method (image 1)]{%
\includegraphics[width=0.38\textwidth]{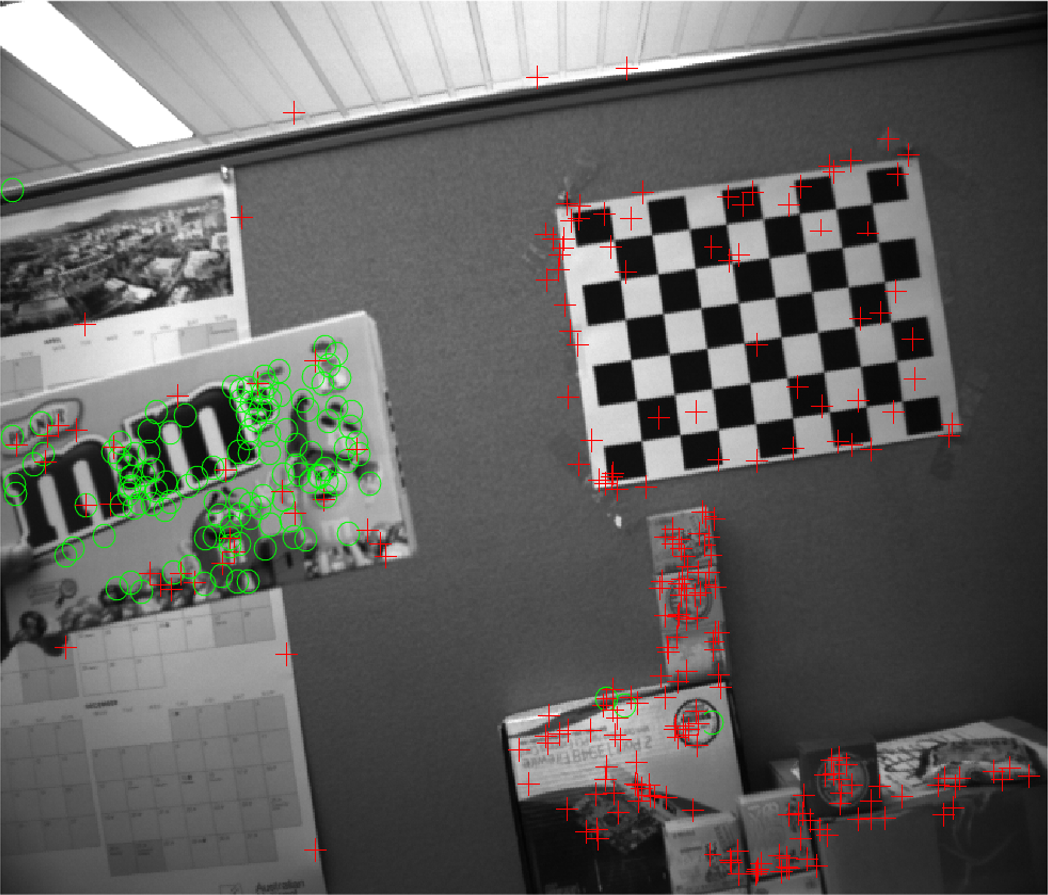}
\label{sys_8pt1}}

\caption{\small {\bf Top row:} The configuration of the non-overlapping two camera rig and one sample image of the left camera. Box shown in red is waved to simulate outliers; {\bf Bottom two rows:} Detected inliers (green circle) and outliers (red cross) by two methods. Note that standard 8-point method fails to detect outliers on the moving box, while our method successfully rejects the outliers. We varied the ransac iterations and thresholds, and in no case had the standard 8-point method succeeded.}
\label{fig:indoorexp}
\label{fig:configure_indoor}
\end{figure}

We mount two PointGrey cameras rigidly on a plane with non-overlapping FOVs in order to validate our methods' performance in real situation with high outlier ratios. We calibrated the multi-camera rig by the hand-eye calibration method in \cite{esquivel17}.  We manually generate outliers by waving a texture-rich box in the scene viewed by one camera, and use a wall-fixed chessboard to compute the vertical vanish point, thus obtain the reference vertical direction as inputs to our method, and compare it with standard 8-point \cite{Normalized-8-point:Hartley-1997} method with RANSAC. The configuration of the two camera rig and one sample image in the left camera are given in fig- \ref{fig:configure_indoor}. We set the Ransac iteration of our method and standard 8-point \cite{Normalized-8-point:Hartley-1997} method to 143 and 1000, respectively. The comparison results are given in fig-\ref{fig:indoorexp}.
\subsubsection{Validate on KITTI Visual Odometry.}
\begin{figure}
\centering
\includegraphics[width=0.48\textwidth]{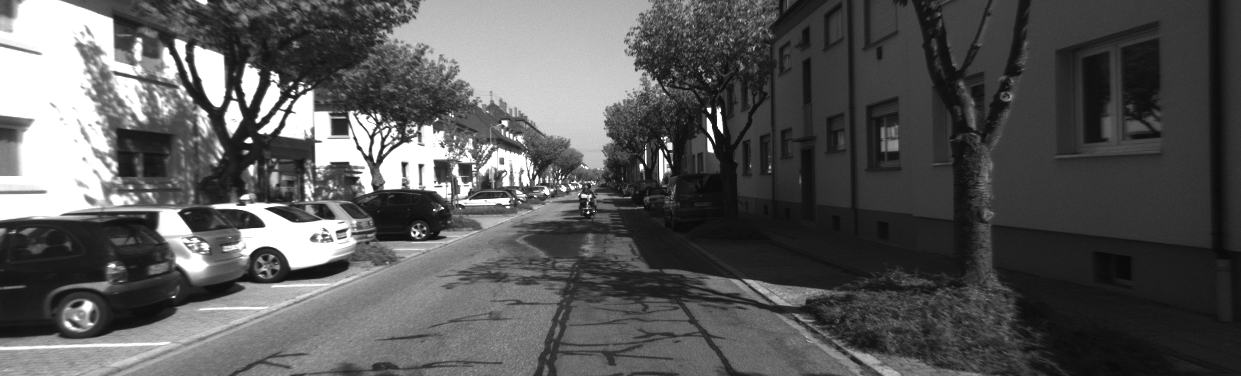}
\includegraphics[width=0.48\textwidth]{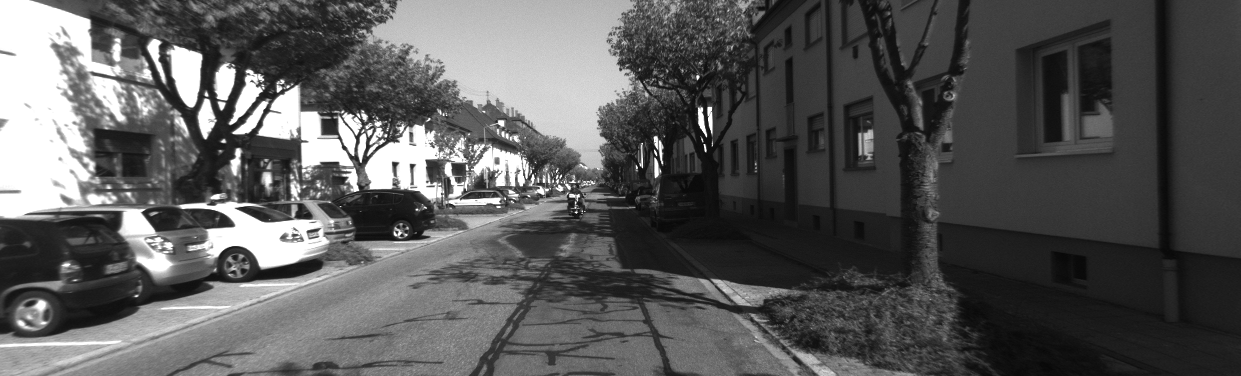}
\caption{\small two sample frames from the left and right camera at frame 20 from the KITTI seq-00 dataset. }
\label{fig:KITTI}
\end{figure}

\begin{figure}[h!]
\centering
\includegraphics[width=0.46\textwidth]{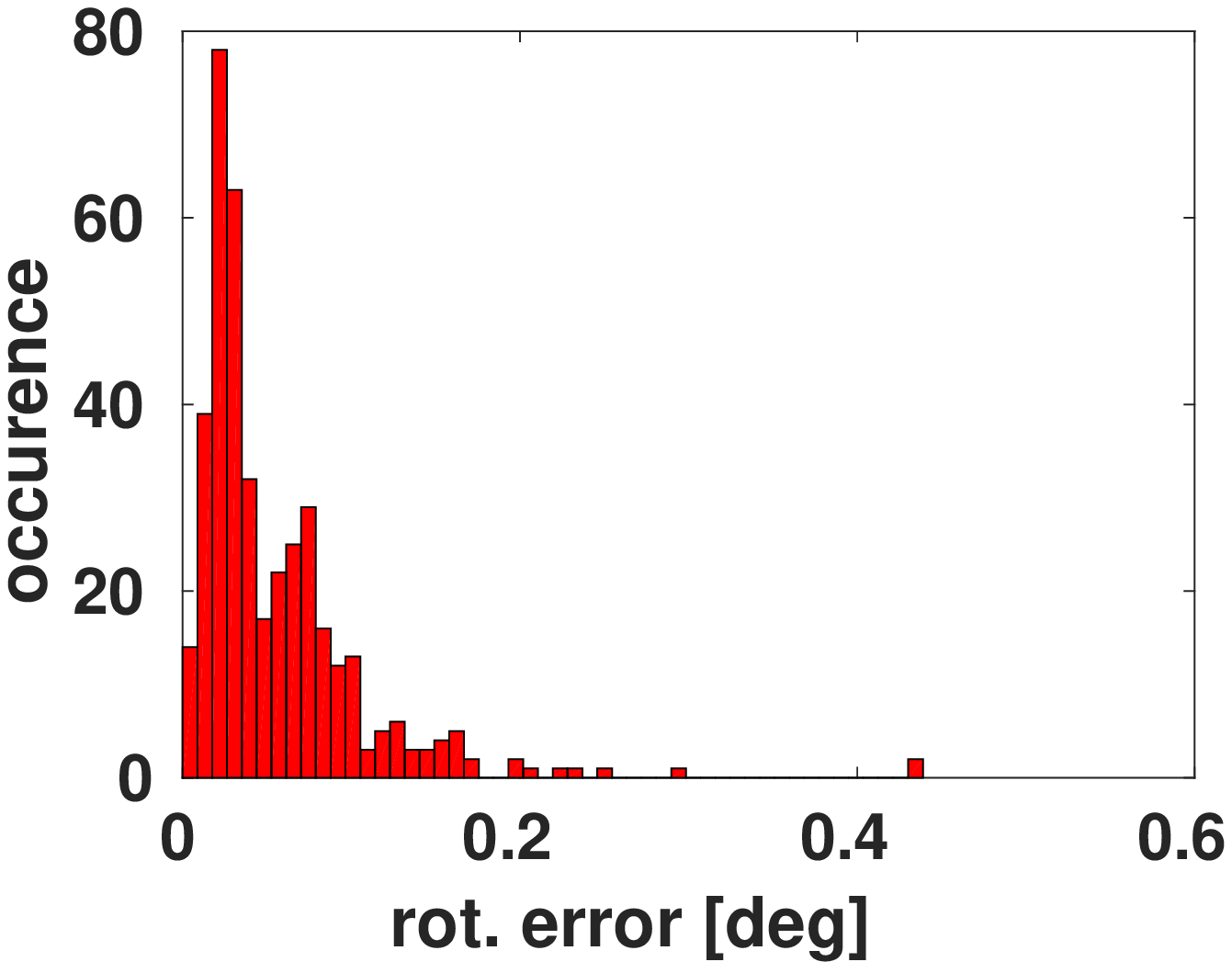}~~~
\includegraphics[width=0.46\textwidth]{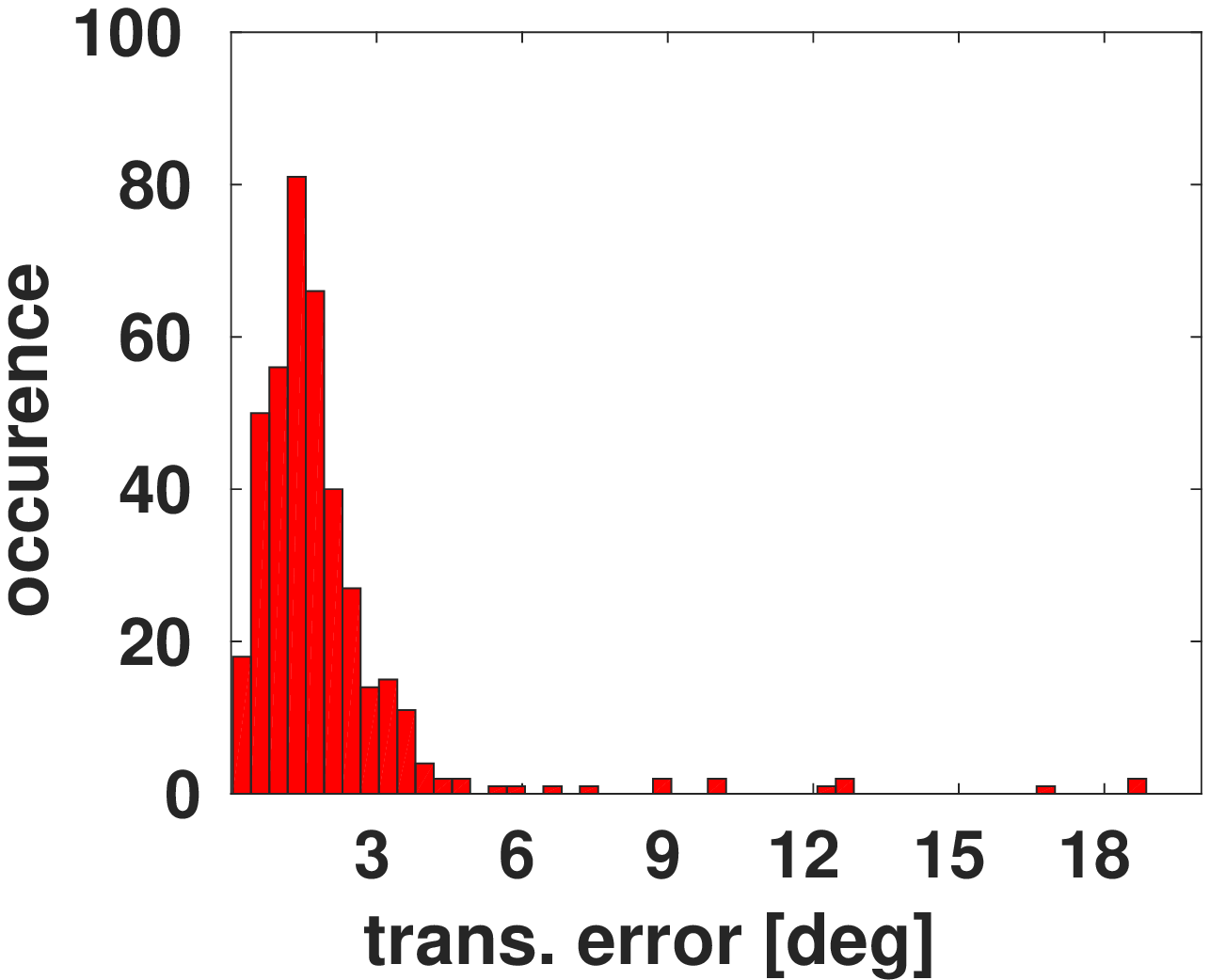}
\caption{The histogram of rotation and translation direction estimation errors for the first 400 frames on KITTI seq-00. Note that most of the rotation errors are under 0.2 degrees and most of the translation direction errors are under 3 degrees.}
\label{fig:KITTI_error}
\end{figure}

We also tested our method on the KITTI autonomous driving benchmark dataset \cite{KITTI-Dataset:CVPR-2012}, despite in KITTI configuration the stereo cameras' FOVs are overlapping.  Since our method uses partial IMU information, it is not our intention (and would not be fair for other methods) to compete on the KITTI Visual Odometry ranking list. The only purpose for this test is to qualitatively verify that our method is applicable to practical driving scenarios.  In our test, feature points are matched between time-consecutive frames in each camera, and we did not match features across the two cameras to mimic a non-overlapping condition.  We fix the number of RANSAC iterations at 143 for our algorithm. Inlier threshold is set at 0.1 degrees in terms of re-projection error. We did not apply inlier set refinement, nor non-linear pose refinement or bundle adjustment, nor no loop closure.  Relative poses from each pair are integrated to get a continuous trajectory of the vehicle. We use the first 400 frames of KITTI seq-00, and two sample images of the sequence are given in figure \ref{fig:KITTI}.  We use the ground-truth roll/pitch angles provided by KITTI, and about 200 matched features per frame are used for the computation. We compare our method with the true rotation and translation directions. As we discussed before, since our method is not able to recover accurate scale, we use the ground truth scale for plotting the trajectory. To have a fair comparison, we also supplied ground-truth scales to Libviso (rather than using Libviso's own scale estimation). We report the error distributions in fig-\ref{fig:KITTI_error}.

\begin{figure}[h!]
\centering
\includegraphics[width=0.65\textwidth]{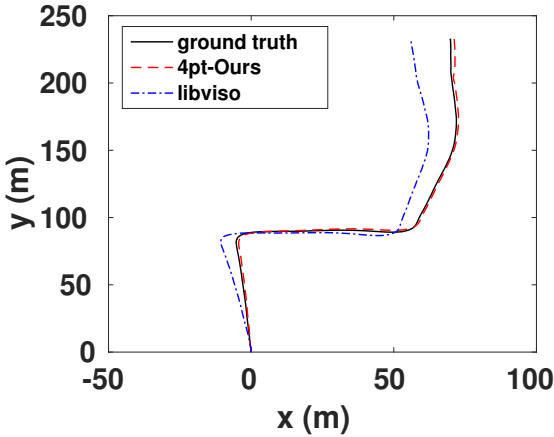}
\caption{\small Estimated Vehicle trajectory (for the first 400 frames in KITTI VO-seq-00); comparing result by our method with the Libviso's monocular VO method \cite{KITTI-Dataset:CVPR-2012}.  For fair comparison, Libviso's scale estimations are replaced with ground-truth scales.}\label{fig:KITTI_traj}
\end{figure}

\section{Conclusions}
Accurately estimating the ego-motion of a vehicle relative to the static background is a key component for autonomous driving and vision-based driving assistance.  In this paper, we have proposed a highly-efficient and highly-robust method for multi-camera ego-motion estimation.  The key idea is to exploit special prior knowledge that are available in the specific application context. Our method is built upon two basic assumptions: small rotation and knowing one reference directional correspondence. Both assumptions are sensible, and are generally available. For example, the directional correspondence may be obtained from either IMU, or a vanishing point, or ground plane fitting.  Moreover, in daily driving scenarios, the rotation between successive video frames (within 20 ms to 100 ms interval) is generally very small. Based on the above assumptions this paper developed an efficient 4-point minimal solver.  A distinct advantage of the method is the use of the conjugate motion idea to efficiently prune out outlier matches in a RANSAC framework. Although the idea is extremely simple and intuitive, experiments show the proposed method is highly efficient in heavily cluttered urban driving environments.
\appendix
\section*{Appendix}

\section{Closed form solution of the quartic equation} \label{closedSOl}
Given a quartic equation $x^4+ax^3+bx^2+cx+d = 0$.  Its four (complex) roots $r_0,r_1,r_2,r_3$ are given by \cite{PlanetMath1}:

\begin{align}
r_0 &= T_1 - R_4 - \sqrt{(R_5 - R_6)}\\
r_1 &= T_1 - R_4 + \sqrt{(R_5 - R_6)}\\
r_2 &= T_1 + R_4 - \sqrt{(R_5 + R_6)}\\
r_3 &= T_1 + R_4 + \sqrt{(R_5 + R_6)}, 
\end{align}where 
\begin{align}
R_1 &= \sqrt{T_3^2 - T_2^3}, \\
R_2 &= \sqrt[3]{T_3 + R_1},\\
R_3 &= (T_2/R_2 + R_2)/12,\\
R_4 &= \sqrt{T_5 + R_3},\\
R_5 &= 2T_5 - R_3,\\
R_6 &= T_4 / R_4,\\
T_1 &= -a/4; \\
T_2 &= b^2 - 3ac + 12d,\\
T_3 &= (2b^3 - 9abc + 27c^2 + 27a^2d - 72bd)/2,\\
T_4 &= (-a^3 + 4ab - 8c)/32,\\
T_5 &= (3a^2 - 8b)/48.
\end{align}

\section{Our C++ source code for 4-point solver}\label{codes}
We list our full source code here to illustrate the simplicity of the algorithm.  Interested reader may also be able to use the code to verify the resulted relative pose. We implemented our 4-point algorithm in C++, making use of the data structures defined in the OpenGV library \cite{OpenGV}.  The OpenGV library is chosen here only for the sake of easy comparison, as it contains many state-of-the-art algorithms for multi-camera relative pose. The expressions for the four coefficients of the quartic equation are generated by Matlab symbolic toolbox. Further simplifications to the expressions may be possible, but we did not explore this due to time reason. The inputs to our function below are four corresponding points (stored in indices) and the known reference direction (stored in adapter).  The output is the relative pose (rotation and translation) represented by a 4x4 Euclidean transformation matrix RTS. The code was tested successfully in Linux/windows environment.
{\footnotesize
\begin{verbatim}
#include <math.h>
#include <complex>
transformations_t fourpt( const RelativeAdapterBase & adapter, 
                  const Indices & indices ){    
  MatrixXd A(4,8);
  Matrix3d rotation = adapter.getR12();
  for( size_t i = 0; i < numberCorrespondences; i++ ){
    bearingVector_t d1 = adapter.getBearingVector1(indices[i]);
    bearingVector_t d2 = adapter.getBearingVector2(indices[i]);
    translation_t v1 = adapter.getCamOffset1(indices[i]);
    translation_t v2 = adapter.getCamOffset2(indices[i]);
    rotation_t R11 = adapter.getCamRotation1(indices[i]);
    rotation_t R21 = adapter.getCamRotation2(indices[i]);
    d1 = R11*d1;
    d2 = R21*d2;
    Eigen::Matrix<double,6,1> l1,l2_pr;
    l1.block<3,1>(0,0) = d1;
    l1.block<3,1>(3,0) = v1.cross(d1);
    l2_pr.block<3,1>(0,0) = rotation*d2;
    l2_pr.block<3,1>(3,0) = rotation*(v2.cross(d2));
    A(i,0) = l1(0)*l2_pr(3) + l1(1)*l2_pr(4) + l1(2)*l2_pr(5) +
             l1(3)*l2_pr(0) + l1(4)*l2_pr(1) + l1(5)*l2_pr(2);
    A(i,1) = l1(2)*l2_pr(1) - l1(1)*l2_pr(2);
    A(i,2) = l1(0)*l2_pr(2) - l1(2)*l2_pr(0);
    A(i,3) = l1(1)*l2_pr(0) - l1(0)*l2_pr(1);
    A(i,4) = l1(1)*l2_pr(3) - l1(3)*l2_pr(1) + l1(4)*l2_pr(0) -
             l1(0)*l2_pr(4);
    A(i,5) = l1(2)*l2_pr(0);
    A(i,6) = l1(2)*l2_pr(1);
    A(i,7) = - l1(0)*l2_pr(0) - l1(1)*l2_pr(1);}
  // Form a quartic equation of x^4+a3*x^3+a2*x^2+a1*x+a0=0 
  double a0 =(-A(0,0)*A(1,1)*A(2,2)*A(3,3)+A(0,0)*A(1,1)*A(2,3)*A(3,2)
  +A(0,0)*A(1,2)*A(2,1)*A(3,3)-A(0,0)*A(1,2)*A(2,3)*A(3,1)-A(0,0)*
  A(1,3)*A(2,1)*A(3,2)+A(0,0)*A(1,3)*A(2,2)*A(3,1)+A(0,1)*A(1,0)*
  A(2,2)*A(3,3)-A(0,1)*A(1,0)*A(2,3)*A(3,2)-A(0,1)*A(1,2)*A(2,0)*
  A(3,3)+A(0,1)*A(1,2)*A(2,3)*A(3,0)+A(0,1)*A(1,3)*A(2,0)*A(3,2)-
  A(0,1)*A(1,3)*A(2,2)*A(3,0)-A(0,2)*A(1,0)*A(2,1)*A(3,3)+A(0,2)*
  A(1,0)*A(2,3)*A(3,1)+A(0,2)*A(1,1)*A(2,0)*A(3,3)-A(0,2)*A(1,1)*
  A(2,3)*A(3,0)-A(0,2)*A(1,3)*A(2,0)*A(3,1)+A(0,2)*A(1,3)*A(2,1)*
  A(3,0)+A(0,3)*A(1,0)*A(2,1)*A(3,2)-A(0,3)*A(1,0)*A(2,2)*A(3,1)-
  A(0,3)*A(1,1)*A(2,0)*A(3,2)+A(0,3)*A(1,1)*A(2,2)*A(3,0)+A(0,3)*
  A(1,2)*A(2,0)*A(3,1)-A(0,3)*A(1,2)*A(2,1)*A(3,0))/(-A(0,4)*A(1,5)
  *A(2,6)*A(3,7)+A(0,4)*A(1,5)*A(2,7)*A(3,6)+A(0,4)*A(1,6)*A(2,5)*
  A(3,7)-A(0,4)*A(1,6)*A(2,7)*A(3,5)-A(0,4)*A(1,7)*A(2,5)*A(3,6)+
  A(0,4)*A(1,7)*A(2,6)*A(3,5)+A(0,5)*A(1,4)*A(2,6)*A(3,7)-A(0,5)*
  A(1,4)*A(2,7)*A(3,6)-A(0,5)*A(1,6)*A(2,4)*A(3,7)+A(0,5)*A(1,6)*
  A(2,7)*A(3,4)+A(0,5)*A(1,7)*A(2,4)*A(3,6)-A(0,5)*A(1,7)*A(2,6)*
  A(3,4)-A(0,6)*A(1,4)*A(2,5)*A(3,7)+A(0,6)*A(1,4)*A(2,7)*A(3,5)+
  A(0,6)*A(1,5)*A(2,4)*A(3,7)-A(0,6)*A(1,5)*A(2,7)*A(3,4)-A(0,6)*
  A(1,7)*A(2,4)*A(3,5)+A(0,6)*A(1,7)*A(2,5)*A(3,4)+A(0,7)*A(1,4)*
  A(2,5)*A(3,6)-A(0,7)*A(1,4)*A(2,6)*A(3,5)-A(0,7)*A(1,5)*A(2,4)*
  A(3,6)+A(0,7)*A(1,5)*A(2,6)*A(3,4)+A(0,7)*A(1,6)*A(2,4)*A(3,5)-
  A(0,7)*A(1,6)*A(2,5)*A(3,4));
  double a1 = (-A(0,0)*A(1,1)*A(2,2)*A(3,7)+A(0,0)*A(1,1)*A(2,3)*
  A(3,6)-A(0,0)*A(1,1)*A(2,6)*A(3,3)+A(0,0)*A(1,1)*A(2,7)*A(3,2)+
  A(0,0)*A(1,2)*A(2,1)*A(3,7)-A(0,0)*A(1,2)*A(2,3)*A(3,5)+A(0,0)*
  A(1,2)*A(2,5)*A(3,3)-A(0,0)*A(1,2)*A(2,7)*A(3,1)-A(0,0)*A(1,3)*
  A(2,1)*A(3,6)+A(0,0)*A(1,3)*A(2,2)*A(3,5)-A(0,0)*A(1,3)*A(2,5)*
  A(3,2)+A(0,0)*A(1,3)*A(2,6)*A(3,1)-A(0,0)*A(1,5)*A(2,2)*A(3,3)+
  A(0,0)*A(1,5)*A(2,3)*A(3,2)+A(0,0)*A(1,6)*A(2,1)*A(3,3)-A(0,0)*
  A(1,6)*A(2,3)*A(3,1)-A(0,0)*A(1,7)*A(2,1)*A(3,2)+A(0,0)*A(1,7)*
  A(2,2)*A(3,1)+A(0,1)*A(1,0)*A(2,2)*A(3,7)-A(0,1)*A(1,0)*A(2,3)*
  A(3,6)+A(0,1)*A(1,0)*A(2,6)*A(3,3)-A(0,1)*A(1,0)*A(2,7)*A(3,2)-
  A(0,1)*A(1,2)*A(2,0)*A(3,7)+A(0,1)*A(1,2)*A(2,3)*A(3,4)-A(0,1)*
  A(1,2)*A(2,4)*A(3,3)+A(0,1)*A(1,2)*A(2,7)*A(3,0)+A(0,1)*A(1,3)*
  A(2,0)*A(3,6)-A(0,1)*A(1,3)*A(2,2)*A(3,4)+A(0,1)*A(1,3)*A(2,4)*
  A(3,2)-A(0,1)*A(1,3)*A(2,6)*A(3,0)+A(0,1)*A(1,4)*A(2,2)*A(3,3)-
  A(0,1)*A(1,4)*A(2,3)*A(3,2)-A(0,1)*A(1,6)*A(2,0)*A(3,3)+A(0,1)*
  A(1,6)*A(2,3)*A(3,0)+A(0,1)*A(1,7)*A(2,0)*A(3,2)-A(0,1)*A(1,7)*
  A(2,2)*A(3,0)-A(0,2)*A(1,0)*A(2,1)*A(3,7)+A(0,2)*A(1,0)*A(2,3)*
  A(3,5)-A(0,2)*A(1,0)*A(2,5)*A(3,3)+A(0,2)*A(1,0)*A(2,7)*A(3,1)+
  A(0,2)*A(1,1)*A(2,0)*A(3,7)-A(0,2)*A(1,1)*A(2,3)*A(3,4)+A(0,2)*
  A(1,1)*A(2,4)*A(3,3)-A(0,2)*A(1,1)*A(2,7)*A(3,0)-A(0,2)*A(1,3)*
  A(2,0)*A(3,5)+A(0,2)*A(1,3)*A(2,1)*A(3,4)-A(0,2)*A(1,3)*A(2,4)*
  A(3,1)+A(0,2)*A(1,3)*A(2,5)*A(3,0)-A(0,2)*A(1,4)*A(2,1)*A(3,3)+
  A(0,2)*A(1,4)*A(2,3)*A(3,1)+A(0,2)*A(1,5)*A(2,0)*A(3,3)-A(0,2)*
  A(1,5)*A(2,3)*A(3,0)-A(0,2)*A(1,7)*A(2,0)*A(3,1)+A(0,2)*A(1,7)*
  A(2,1)*A(3,0)+A(0,3)*A(1,0)*A(2,1)*A(3,6)-A(0,3)*A(1,0)*A(2,2)*
  A(3,5)+A(0,3)*A(1,0)*A(2,5)*A(3,2)-A(0,3)*A(1,0)*A(2,6)*A(3,1)-
  A(0,3)*A(1,1)*A(2,0)*A(3,6)+A(0,3)*A(1,1)*A(2,2)*A(3,4)-A(0,3)*
  A(1,1)*A(2,4)*A(3,2)+A(0,3)*A(1,1)*A(2,6)*A(3,0)+A(0,3)*A(1,2)*
  A(2,0)*A(3,5)-A(0,3)*A(1,2)*A(2,1)*A(3,4)+A(0,3)*A(1,2)*A(2,4)*
  A(3,1)-A(0,3)*A(1,2)*A(2,5)*A(3,0)+A(0,3)*A(1,4)*A(2,1)*A(3,2)-
  A(0,3)*A(1,4)*A(2,2)*A(3,1)-A(0,3)*A(1,5)*A(2,0)*A(3,2)+A(0,3)*
  A(1,5)*A(2,2)*A(3,0)+A(0,3)*A(1,6)*A(2,0)*A(3,1)-A(0,3)*A(1,6)*
  A(2,1)*A(3,0)-A(0,4)*A(1,1)*A(2,2)*A(3,3)+A(0,4)*A(1,1)*A(2,3)*
  A(3,2)+A(0,4)*A(1,2)*A(2,1)*A(3,3)-A(0,4)*A(1,2)*A(2,3)*A(3,1)-
  A(0,4)*A(1,3)*A(2,1)*A(3,2)+A(0,4)*A(1,3)*A(2,2)*A(3,1)+A(0,5)*
  A(1,0)*A(2,2)*A(3,3)-A(0,5)*A(1,0)*A(2,3)*A(3,2)-A(0,5)*A(1,2)*
  A(2,0)*A(3,3)+A(0,5)*A(1,2)*A(2,3)*A(3,0)+A(0,5)*A(1,3)*A(2,0)*
  A(3,2)-A(0,5)*A(1,3)*A(2,2)*A(3,0)-A(0,6)*A(1,0)*A(2,1)*A(3,3)+
  A(0,6)*A(1,0)*A(2,3)*A(3,1)+A(0,6)*A(1,1)*A(2,0)*A(3,3)-A(0,6)*
  A(1,1)*A(2,3)*A(3,0)-A(0,6)*A(1,3)*A(2,0)*A(3,1)+A(0,6)*A(1,3)*
  A(2,1)*A(3,0)+A(0,7)*A(1,0)*A(2,1)*A(3,2)-A(0,7)*A(1,0)*A(2,2)*
  A(3,1)-A(0,7)*A(1,1)*A(2,0)*A(3,2)+A(0,7)*A(1,1)*A(2,2)*A(3,0)+
  A(0,7)*A(1,2)*A(2,0)*A(3,1)-A(0,7)*A(1,2)*A(2,1)*A(3,0))/(-A(0,4)
  *A(1,5)*A(2,6)*A(3,7)+A(0,4)*A(1,5)*A(2,7)*A(3,6)+A(0,4)*A(1,6)
  *A(2,5)*A(3,7)-A(0,4)*A(1,6)*A(2,7)*A(3,5)-A(0,4)*A(1,7)*A(2,5)*
  A(3,6)+A(0,4)*A(1,7)*A(2,6)*A(3,5)+A(0,5)*A(1,4)*A(2,6)*A(3,7)-
  A(0,5)*A(1,4)*A(2,7)*A(3,6)-A(0,5)*A(1,6)*A(2,4)*A(3,7)+A(0,5)*
  A(1,6)*A(2,7)*A(3,4)+A(0,5)*A(1,7)*A(2,4)*A(3,6)-A(0,5)*A(1,7)*
  A(2,6)*A(3,4)-A(0,6)*A(1,4)*A(2,5)*A(3,7)+A(0,6)*A(1,4)*A(2,7)*
  A(3,5)+A(0,6)*A(1,5)*A(2,4)*A(3,7)-A(0,6)*A(1,5)*A(2,7)*A(3,4)
  -A(0,6)*A(1,7)*A(2,4)*A(3,5)+A(0,6)*A(1,7)*A(2,5)*A(3,4)+A(0,7)*
  A(1,4)*A(2,5)*A(3,6)-A(0,7)*A(1,4)*A(2,6)*A(3,5)-A(0,7)*A(1,5)*
  A(2,4)*A(3,6)+A(0,7)*A(1,5)*A(2,6)*A(3,4)+A(0,7)*A(1,6)*A(2,4)*
  A(3,5)-A(0,7)*A(1,6)*A(2,5)*A(3,4));
  double a2 = (-A(0,0)*A(1,1)*A(2,6)*A(3,7)+A(0,0)*A(1,1)*A(2,7)*
  A(3,6)+A(0,0)*A(1,2)*A(2,5)*A(3,7)-A(0,0)*A(1,2)*A(2,7)*A(3,5)
  -A(0,0)*A(1,3)*A(2,5)*A(3,6)+A(0,0)*A(1,3)*A(2,6)*A(3,5)-A(0,0)*
  A(1,5)*A(2,2)*A(3,7)+A(0,0)*A(1,5)*A(2,3)*A(3,6)-A(0,0)*A(1,5)*
  A(2,6)*A(3,3)+A(0,0)*A(1,5)*A(2,7)*A(3,2)+A(0,0)*A(1,6)*A(2,1)*
  A(3,7)-A(0,0)*A(1,6)*A(2,3)*A(3,5)+A(0,0)*A(1,6)*A(2,5)*A(3,3)-
  A(0,0)*A(1,6)*A(2,7)*A(3,1)-A(0,0)*A(1,7)*A(2,1)*A(3,6)+A(0,0)*
  A(1,7)*A(2,2)*A(3,5)-A(0,0)*A(1,7)*A(2,5)*A(3,2)+A(0,0)*A(1,7)*
  A(2,6)*A(3,1)+A(0,1)*A(1,0)*A(2,6)*A(3,7)-A(0,1)*A(1,0)*A(2,7)*
  A(3,6)-A(0,1)*A(1,2)*A(2,4)*A(3,7)+A(0,1)*A(1,2)*A(2,7)*A(3,4)+
  A(0,1)*A(1,3)*A(2,4)*A(3,6)-A(0,1)*A(1,3)*A(2,6)*A(3,4)+A(0,1)*
  A(1,4)*A(2,2)*A(3,7)-A(0,1)*A(1,4)*A(2,3)*A(3,6)+A(0,1)*A(1,4)*
  A(2,6)*A(3,3)-A(0,1)*A(1,4)*A(2,7)*A(3,2)-A(0,1)*A(1,6)*A(2,0)*
  A(3,7)+A(0,1)*A(1,6)*A(2,3)*A(3,4)-A(0,1)*A(1,6)*A(2,4)*A(3,3)+
  A(0,1)*A(1,6)*A(2,7)*A(3,0)+A(0,1)*A(1,7)*A(2,0)*A(3,6)-A(0,1)*
  A(1,7)*A(2,2)*A(3,4)+A(0,1)*A(1,7)*A(2,4)*A(3,2)-A(0,1)*A(1,7)*
  A(2,6)*A(3,0)-A(0,2)*A(1,0)*A(2,5)*A(3,7)+A(0,2)*A(1,0)*A(2,7)*
  A(3,5)+A(0,2)*A(1,1)*A(2,4)*A(3,7)-A(0,2)*A(1,1)*A(2,7)*A(3,4)-
  A(0,2)*A(1,3)*A(2,4)*A(3,5)+A(0,2)*A(1,3)*A(2,5)*A(3,4)-A(0,2)*
  A(1,4)*A(2,1)*A(3,7)+A(0,2)*A(1,4)*A(2,3)*A(3,5)-A(0,2)*A(1,4)*
  A(2,5)*A(3,3)+A(0,2)*A(1,4)*A(2,7)*A(3,1)+A(0,2)*A(1,5)*A(2,0)*
  A(3,7)-A(0,2)*A(1,5)*A(2,3)*A(3,4)+A(0,2)*A(1,5)*A(2,4)*A(3,3)-
  A(0,2)*A(1,5)*A(2,7)*A(3,0)-A(0,2)*A(1,7)*A(2,0)*A(3,5)+A(0,2)*
  A(1,7)*A(2,1)*A(3,4)-A(0,2)*A(1,7)*A(2,4)*A(3,1)+A(0,2)*A(1,7)*
  A(2,5)*A(3,0)+A(0,3)*A(1,0)*A(2,5)*A(3,6)-A(0,3)*A(1,0)*A(2,6)*
  A(3,5)-A(0,3)*A(1,1)*A(2,4)*A(3,6)+A(0,3)*A(1,1)*A(2,6)*A(3,4)+
  A(0,3)*A(1,2)*A(2,4)*A(3,5)-A(0,3)*A(1,2)*A(2,5)*A(3,4)+A(0,3)*
  A(1,4)*A(2,1)*A(3,6)-A(0,3)*A(1,4)*A(2,2)*A(3,5)+A(0,3)*A(1,4)*
  A(2,5)*A(3,2)-A(0,3)*A(1,4)*A(2,6)*A(3,1)-A(0,3)*A(1,5)*A(2,0)*
  A(3,6)+A(0,3)*A(1,5)*A(2,2)*A(3,4)-A(0,3)*A(1,5)*A(2,4)*A(3,2)+
  A(0,3)*A(1,5)*A(2,6)*A(3,0)+A(0,3)*A(1,6)*A(2,0)*A(3,5)-A(0,3)*
  A(1,6)*A(2,1)*A(3,4)+A(0,3)*A(1,6)*A(2,4)*A(3,1)-A(0,3)*A(1,6)*
  A(2,5)*A(3,0)-A(0,4)*A(1,1)*A(2,2)*A(3,7)+A(0,4)*A(1,1)*A(2,3)*
  A(3,6)-A(0,4)*A(1,1)*A(2,6)*A(3,3)+A(0,4)*A(1,1)*A(2,7)*A(3,2)+
  A(0,4)*A(1,2)*A(2,1)*A(3,7)-A(0,4)*A(1,2)*A(2,3)*A(3,5)+A(0,4)*
  A(1,2)*A(2,5)*A(3,3)-A(0,4)*A(1,2)*A(2,7)*A(3,1)-A(0,4)*A(1,3)*
  A(2,1)*A(3,6)+A(0,4)*A(1,3)*A(2,2)*A(3,5)-A(0,4)*A(1,3)*A(2,5)*
  A(3,2)+A(0,4)*A(1,3)*A(2,6)*A(3,1)-A(0,4)*A(1,5)*A(2,2)*A(3,3)+
  A(0,4)*A(1,5)*A(2,3)*A(3,2)+A(0,4)*A(1,6)*A(2,1)*A(3,3)-A(0,4)*
  A(1,6)*A(2,3)*A(3,1)-A(0,4)*A(1,7)*A(2,1)*A(3,2)+A(0,4)*A(1,7)*
  A(2,2)*A(3,1)+A(0,5)*A(1,0)*A(2,2)*A(3,7)-A(0,5)*A(1,0)*A(2,3)*
  A(3,6)+A(0,5)*A(1,0)*A(2,6)*A(3,3)-A(0,5)*A(1,0)*A(2,7)*A(3,2)
  -A(0,5)*A(1,2)*A(2,0)*A(3,7)+A(0,5)*A(1,2)*A(2,3)*A(3,4)-A(0,5)*
  A(1,2)*A(2,4)*A(3,3)+A(0,5)*A(1,2)*A(2,7)*A(3,0)+A(0,5)*A(1,3)*
  A(2,0)*A(3,6)-A(0,5)*A(1,3)*A(2,2)*A(3,4)+A(0,5)*A(1,3)*A(2,4)*
  A(3,2)-A(0,5)*A(1,3)*A(2,6)*A(3,0)+A(0,5)*A(1,4)*A(2,2)*A(3,3)-
  A(0,5)*A(1,4)*A(2,3)*A(3,2)-A(0,5)*A(1,6)*A(2,0)*A(3,3)+A(0,5)*
  A(1,6)*A(2,3)*A(3,0)+A(0,5)*A(1,7)*A(2,0)*A(3,2)-A(0,5)*A(1,7)*
  A(2,2)*A(3,0)-A(0,6)*A(1,0)*A(2,1)*A(3,7)+A(0,6)*A(1,0)*A(2,3)*
  A(3,5)-A(0,6)*A(1,0)*A(2,5)*A(3,3)+A(0,6)*A(1,0)*A(2,7)*A(3,1)+
  A(0,6)*A(1,1)*A(2,0)*A(3,7)-A(0,6)*A(1,1)*A(2,3)*A(3,4)+A(0,6)*
  A(1,1)*A(2,4)*A(3,3)-A(0,6)*A(1,1)*A(2,7)*A(3,0)-A(0,6)*A(1,3)*
  A(2,0)*A(3,5)+A(0,6)*A(1,3)*A(2,1)*A(3,4)-A(0,6)*A(1,3)*A(2,4)*
  A(3,1)+A(0,6)*A(1,3)*A(2,5)*A(3,0)-A(0,6)*A(1,4)*A(2,1)*A(3,3)+
  A(0,6)*A(1,4)*A(2,3)*A(3,1)+A(0,6)*A(1,5)*A(2,0)*A(3,3)-A(0,6)*
  A(1,5)*A(2,3)*A(3,0)-A(0,6)*A(1,7)*A(2,0)*A(3,1)+A(0,6)*A(1,7)*
  A(2,1)*A(3,0)+A(0,7)*A(1,0)*A(2,1)*A(3,6)-A(0,7)*A(1,0)*A(2,2)*
  A(3,5)+A(0,7)*A(1,0)*A(2,5)*A(3,2)-A(0,7)*A(1,0)*A(2,6)*A(3,1)
  -A(0,7)*A(1,1)*A(2,0)*A(3,6)+A(0,7)*A(1,1)*A(2,2)*A(3,4)-A(0,7)*
  A(1,1)*A(2,4)*A(3,2)+A(0,7)*A(1,1)*A(2,6)*A(3,0)+A(0,7)*A(1,2)*
  A(2,0)*A(3,5)-A(0,7)*A(1,2)*A(2,1)*A(3,4)+A(0,7)*A(1,2)*A(2,4)*
  A(3,1)-A(0,7)*A(1,2)*A(2,5)*A(3,0)+A(0,7)*A(1,4)*A(2,1)*A(3,2)-
  A(0,7)*A(1,4)*A(2,2)*A(3,1)-A(0,7)*A(1,5)*A(2,0)*A(3,2)+A(0,7)*
  A(1,5)*A(2,2)*A(3,0)+A(0,7)*A(1,6)*A(2,0)*A(3,1)-A(0,7)*A(1,6)*
  A(2,1)*A(3,0))/(-A(0,4)*A(1,5)*A(2,6)*A(3,7)+A(0,4)*A(1,5)*
  A(2,7)*A(3,6)+A(0,4)*A(1,6)*A(2,5)*A(3,7)-A(0,4)*A(1,6)*A(2,7)*
  A(3,5)-A(0,4)*A(1,7)*A(2,5)*A(3,6)+A(0,4)*A(1,7)*A(2,6)*A(3,5)+
  A(0,5)*A(1,4)*A(2,6)*A(3,7)-A(0,5)*A(1,4)*A(2,7)*A(3,6)
  -A(0,5)*A(1,6)*A(2,4)*A(3,7)+A(0,5)*A(1,6)*A(2,7)*A(3,4)+A(0,5)*
  A(1,7)*A(2,4)*A(3,6)-A(0,5)*A(1,7)*A(2,6)*A(3,4)-A(0,6)*A(1,4)*
  A(2,5)*A(3,7)+A(0,6)*A(1,4)*A(2,7)*A(3,5)+A(0,6)*A(1,5)*A(2,4)*
  A(3,7)-A(0,6)*A(1,5)*A(2,7)*A(3,4)-A(0,6)*A(1,7)*A(2,4)*A(3,5)+
  A(0,6)*A(1,7)*A(2,5)*A(3,4)+A(0,7)*A(1,4)*A(2,5)*A(3,6)-A(0,7)*
  A(1,4)*A(2,6)*A(3,5)-A(0,7)*A(1,5)*A(2,4)*A(3,6)+A(0,7)*A(1,5)*
  A(2,6)*A(3,4)+A(0,7)*A(1,6)*A(2,4)*A(3,5)-A(0,7)*A(1,6)*A(2,5)*
  A(3,4));
  double a3 = (-A(0,0)*A(1,5)*A(2,6)*A(3,7)+A(0,0)*A(1,5)*A(2,7)*
  A(3,6)+A(0,0)*A(1,6)*A(2,5)*A(3,7)-A(0,0)*A(1,6)*A(2,7)*A(3,5)
  -A(0,0)*A(1,7)*A(2,5)*A(3,6)+A(0,0)*A(1,7)*A(2,6)*A(3,5)+A(0,1)
  *A(1,4)*A(2,6)*A(3,7)-A(0,1)*A(1,4)*A(2,7)*A(3,6)-A(0,1)*A(1,6)*
  A(2,4)*A(3,7)+A(0,1)*A(1,6)*A(2,7)*A(3,4)+A(0,1)*A(1,7)*A(2,4)*
  A(3,6)-A(0,1)*A(1,7)*A(2,6)*A(3,4)-A(0,2)*A(1,4)*A(2,5)*A(3,7)+
  A(0,2)*A(1,4)*A(2,7)*A(3,5)+A(0,2)*A(1,5)*A(2,4)*A(3,7)-A(0,2)*
  A(1,5)*A(2,7)*A(3,4)-A(0,2)*A(1,7)*A(2,4)*A(3,5)+A(0,2)*A(1,7)*
  A(2,5)*A(3,4)+A(0,3)*A(1,4)*A(2,5)*A(3,6)-A(0,3)*A(1,4)*A(2,6)*
  A(3,5)-A(0,3)*A(1,5)*A(2,4)*A(3,6)+A(0,3)*A(1,5)*A(2,6)*A(3,4)+
  A(0,3)*A(1,6)*A(2,4)*A(3,5)-A(0,3)*A(1,6)*A(2,5)*A(3,4)-A(0,4)*
  A(1,1)*A(2,6)*A(3,7)+A(0,4)*A(1,1)*A(2,7)*A(3,6)+A(0,4)*A(1,2)*
  A(2,5)*A(3,7)-A(0,4)*A(1,2)*A(2,7)*A(3,5)-A(0,4)*A(1,3)*A(2,5)*
  A(3,6)+A(0,4)*A(1,3)*A(2,6)*A(3,5)-A(0,4)*A(1,5)*A(2,2)*A(3,7)+
  A(0,4)*A(1,5)*A(2,3)*A(3,6)-A(0,4)*A(1,5)*A(2,6)*A(3,3)+A(0,4)*
  A(1,5)*A(2,7)*A(3,2)+A(0,4)*A(1,6)*A(2,1)*A(3,7)-A(0,4)*A(1,6)*
  A(2,3)*A(3,5)+A(0,4)*A(1,6)*A(2,5)*A(3,3)-A(0,4)*A(1,6)*A(2,7)*
  A(3,1)-A(0,4)*A(1,7)*A(2,1)*A(3,6)+A(0,4)*A(1,7)*A(2,2)*A(3,5)
  -A(0,4)*A(1,7)*A(2,5)*A(3,2)+A(0,4)*A(1,7)*A(2,6)*A(3,1)+A(0,5)*
  A(1,0)*A(2,6)*A(3,7)-A(0,5)*A(1,0)*A(2,7)*A(3,6)-A(0,5)*A(1,2)*
  A(2,4)*A(3,7)+A(0,5)*A(1,2)*A(2,7)*A(3,4)+A(0,5)*A(1,3)*A(2,4)*
  A(3,6)-A(0,5)*A(1,3)*A(2,6)*A(3,4)+A(0,5)*A(1,4)*A(2,2)*A(3,7)
  -A(0,5)*A(1,4)*A(2,3)*A(3,6)+A(0,5)*A(1,4)*A(2,6)*A(3,3)-A(0,5)*
  A(1,4)*A(2,7)*A(3,2)-A(0,5)*A(1,6)*A(2,0)*A(3,7)+A(0,5)*A(1,6)*
  A(2,3)*A(3,4)-A(0,5)*A(1,6)*A(2,4)*A(3,3)+A(0,5)*A(1,6)*A(2,7)*
  A(3,0)+A(0,5)*A(1,7)*A(2,0)*A(3,6)-A(0,5)*A(1,7)*A(2,2)*A(3,4)+
  A(0,5)*A(1,7)*A(2,4)*A(3,2)-A(0,5)*A(1,7)*A(2,6)*A(3,0)-A(0,6)*
  A(1,0)*A(2,5)*A(3,7)+A(0,6)*A(1,0)*A(2,7)*A(3,5)+A(0,6)*A(1,1)*
  A(2,4)*A(3,7)-A(0,6)*A(1,1)*A(2,7)*A(3,4)-A(0,6)*A(1,3)*A(2,4)*
  A(3,5)+A(0,6)*A(1,3)*A(2,5)*A(3,4)-A(0,6)*A(1,4)*A(2,1)*A(3,7)+
  A(0,6)*A(1,4)*A(2,3)*A(3,5)-A(0,6)*A(1,4)*A(2,5)*A(3,3)+A(0,6)*
  A(1,4)*A(2,7)*A(3,1)+A(0,6)*A(1,5)*A(2,0)*A(3,7)-A(0,6)*A(1,5)*
  A(2,3)*A(3,4)+A(0,6)*A(1,5)*A(2,4)*A(3,3)-A(0,6)*A(1,5)*A(2,7)*
  A(3,0)-A(0,6)*A(1,7)*A(2,0)*A(3,5)+A(0,6)*A(1,7)*A(2,1)*A(3,4)
  -A(0,6)*A(1,7)*A(2,4)*A(3,1)+A(0,6)*A(1,7)*A(2,5)*A(3,0)+A(0,7)*
  A(1,0)*A(2,5)*A(3,6)-A(0,7)*A(1,0)*A(2,6)*A(3,5)-A(0,7)*A(1,1)*
  A(2,4)*A(3,6)+A(0,7)*A(1,1)*A(2,6)*A(3,4)+A(0,7)*A(1,2)*A(2,4)*
  A(3,5)-A(0,7)*A(1,2)*A(2,5)*A(3,4)+A(0,7)*A(1,4)*A(2,1)*A(3,6)
  -A(0,7)*A(1,4)*A(2,2)*A(3,5)+A(0,7)*A(1,4)*A(2,5)*A(3,2)-A(0,7)*
  A(1,4)*A(2,6)*A(3,1)-A(0,7)*A(1,5)*A(2,0)*A(3,6)+A(0,7)*A(1,5)*
  A(2,2)*A(3,4)-A(0,7)*A(1,5)*A(2,4)*A(3,2)+A(0,7)*A(1,5)*A(2,6)*
  A(3,0)+A(0,7)*A(1,6)*A(2,0)*A(3,5)-A(0,7)*A(1,6)*A(2,1)*A(3,4)+
  A(0,7)*A(1,6)*A(2,4)*A(3,1)-A(0,7)*A(1,6)*A(2,5)*A(3,0))/(-A(0,4)
  *A(1,5)*A(2,6)*A(3,7)+A(0,4)*A(1,5)*A(2,7)*A(3,6)+A(0,4)*A(1,6)*
  A(2,5)*A(3,7)-A(0,4)*A(1,6)*A(2,7)*A(3,5)-A(0,4)*A(1,7)*A(2,5)*
  A(3,6)+A(0,4)*A(1,7)*A(2,6)*A(3,5)+A(0,5)*A(1,4)*A(2,6)*A(3,7)-
  A(0,5)*A(1,4)*A(2,7)*A(3,6)-A(0,5)*A(1,6)*A(2,4)*A(3,7)+A(0,5)*
  A(1,6)*A(2,7)*A(3,4)+A(0,5)*A(1,7)*A(2,4)*A(3,6)-A(0,5)*A(1,7)*
  A(2,6)*A(3,4)-A(0,6)*A(1,4)*A(2,5)*A(3,7)+A(0,6)*A(1,4)*A(2,7)*
  A(3,5)+A(0,6)*A(1,5)*A(2,4)*A(3,7)-A(0,6)*A(1,5)*A(2,7)*A(3,4)
  -A(0,6)*A(1,7)*A(2,4)*A(3,5)+A(0,6)*A(1,7)*A(2,5)*A(3,4)+A(0,7)*
  A(1,4)*A(2,5)*A(3,6)-A(0,7)*A(1,4)*A(2,6)*A(3,5)-A(0,7)*A(1,5)*
  A(2,4)*A(3,6)+A(0,7)*A(1,5)*A(2,6)*A(3,4)+A(0,7)*A(1,6)*A(2,4)*
  A(3,5)-A(0,7)*A(1,6)*A(2,5)*A(3,4));
  // Solve the quartic equation in closed-form  
  complex<double> T1 = -a3/4.0;
  complex<double> T2 = a2*a2 - 3.0*a3*a1 + 12.0*a0;
  complex<double> T3 = (2.0*a2*a2*a2 - 9.0*a3*a2*a1 + 27.0*a1*a1 +
                       27.0*a3*a3*a0 - 72.0*a2*a0)/2.0;
  complex<double> T4 = (-a3*a3*a3 + 4.0*a3*a2 - 8.0*a1)/32.0;
  complex<double> T5 = (3.0*a3*a3 - 8.0*a2)/48.0;
  complex<double> R1 = sqrt(T3*T3 - T2*T2*T2);
  complex<double> R2 = pow(T3R1, 1.0/3.0);
  complex<double> R3 = (1.0/12.0)*(T2/R2 + R2);
  complex<double> R4 = sqrt(T5 + R3);
  complex<double> R5 = 2.0 *T5 - R3;
  complex<double> R6 = T4 / R4;
  if ((T4 == 0.0) && (T5 == 0.0) && (abs(R3) < 1e-16))
    R6 = 1.0;
  complex<double> r[4];
  r[0] = T1 - R4 - sqrt(R5 - R6);
  r[1] = T1 - R4 + sqrt(R5 - R6);
  r[2] = T1 + R4 - sqrt(R5 + R6);
  r[3] = T1 + R4 + sqrt(R5 + R6);
  std::vector <double> real_roots;
  for (int i = 0; i < 4; ++i){
    if ((abs(r[i].imag()) < 1e-6) && (abs(r[i].real()) < 0.2618)){
    real_roots.push_back(r[i].real());}}
  transformations_t RTS;
  RTS.resize(real_roots.size());
  for (int i = 0; i < real_roots.size(); ++i){
    Matrix3d skew, yaw_m;
    skew << 0.0, -real_roots[i], 0.0,
    real_roots[i], 0.0, 0.0,
    0.0 ,0.0, 0.0;
  if (real_roots[i] == 0.0 ){
    yaw_m = Matrix3d::Identity();}
  else{
    yaw_m = Matrix3d::Identity() + (sin(abs(real_roots[i]))/
    abs(real_roots[i])) * skew + ((1- cos(abs(real_roots[i])))/
    (real_roots[i] * real_roots[i])) * skew * skew;}
  Matrix<double,4,3> M_c;
  Vector4d b_c;
  for (int j = 0; j < 4; ++j){
  M_c(j,0) = A(j,1) + real_roots[i] * A(j,5);
  M_c(j,1) = A(j,2) + real_roots[i] * A(j,6);
  M_c(j,2) = A(j,3) + real_roots[i] * A(j,7);
  b_c(j) = -(A(j,0) + real_roots[i] * A(j,4));}
  Vector3d t = M_c.householderQr().solve(b_c);
  Matrix3d R_est = yaw_m * rotation;
  transformation_t RT;
  RT.block<3,3>(0,0) = R_est;
  RT.block<3,1>(0,3) = t;
  RTS.at(i) = RT;
  }
  return RTS;}
\end{verbatim}
}

{\footnotesize\bibliographystyle{splncs}
\bibliography{Four_Point_Relative_Pose_Reference}} 
\end{document}